\documentclass[10pt,twocolumn,letterpaper]{article}

\usepackage{iccv}
\usepackage{times}
\usepackage{epsfig}
\usepackage{graphicx}
\usepackage{amsmath}
\usepackage{amssymb}
\usepackage{booktabs}
\usepackage{multirow}
\usepackage{enumitem}
\usepackage{color,soul}
\usepackage{relsize}

\usepackage[pagebackref=true,breaklinks=true,letterpaper=true,colorlinks,bookmarks=false]{hyperref}
\usepackage[capitalize]{cleveref}
\crefname{section}{Sec.}{Secs.}
\Crefname{section}{Section}{Sections}
\Crefname{table}{Table}{Tables}
\crefname{table}{Tab.}{Tabs.}

\usepackage{xcolor,colortbl}
\usepackage{siunitx}
\usepackage{amsmath}

\DeclareMathOperator*{\argmin}{argmin}

\newcommand\blfootnote[1]{
  \begingroup
  \renewcommand\thefootnote{}\footnote{#1}
  \addtocounter{footnote}{-1}
  \endgroup
}

\newcommand{\SO}{\ensuremath{\mathbf{SO}}}
\newcommand{\SE}{\ensuremath{\mathbf{SE}}}
\newcommand{\fun}[3]{\ensuremath{#1\colon #2\mapsto #3}}
\newcommand{\R}{\mathbb{R}}

\iccvfinalcopy

\ificcvfinal\fi

\begin{document}

\title{LU-NeRF: Scene and Pose Estimation by Synchronizing Local Unposed NeRFs}

\author{Zezhou Cheng$^{1,2}$\thanks{Work done during an internship at Google.} \quad Carlos Esteves$^2$ \quad Varun Jampani$^2$ \quad Abhishek Kar$^2$ \\ 
Subhransu Maji$^1$ \quad Ameesh Makadia$^2$\\
$^1$University of Massachusetts, Amherst \quad $^2$Google Research\\
}

\maketitle
\ificcvfinal\thispagestyle{empty}\fi

\begin{abstract}
A critical obstacle preventing NeRF models from being deployed broadly in the wild is their reliance on accurate camera poses. Consequently, there is growing interest in extending NeRF models to jointly optimize camera poses and scene representation, which offers an alternative to off-the-shelf SfM pipelines which have well-understood failure modes. Existing approaches for unposed NeRF operate under limiting assumptions, such as a prior pose distribution or coarse pose initialization, making them less effective in a general setting. In this work, we propose a novel approach, LU-NeRF, that jointly estimates camera poses and neural radiance fields with relaxed assumptions on pose configuration.
Our approach operates in a local-to-global manner, where we first optimize over local subsets of the data, dubbed ``mini-scenes.’’ LU-NeRF estimates local pose and geometry for this challenging few-shot task. The mini-scene poses are brought into a global reference frame through a robust pose synchronization step, where a final global optimization of pose and scene can be performed.
We show our LU-NeRF pipeline outperforms prior attempts at unposed NeRF without making restrictive assumptions on the pose prior. This allows us to operate in the general SE(3) pose setting, unlike the baselines. Our results also indicate our model can be complementary to feature-based SfM pipelines as it compares favorably to COLMAP on low-texture and low-resolution images. \blfootnote{\href{http://people.cs.umass.edu/~zezhoucheng/lu-nerf/}{Project website}}
\end{abstract}

\begin{figure*}
\centering
\includegraphics[width=\linewidth]{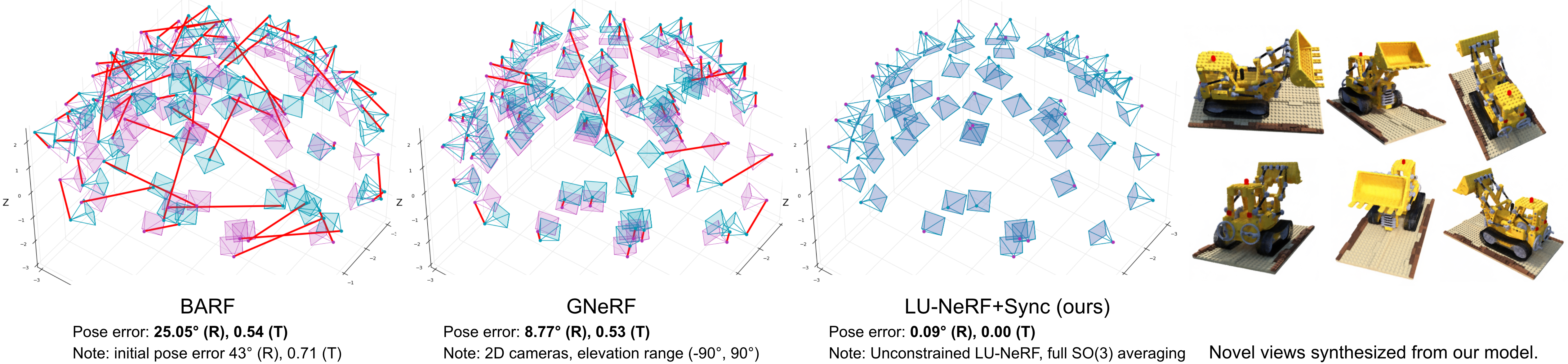}
\caption{Jointly optimizing camera poses and scene representation over a full scene is difficult and underconstrained.
This example is the Lego scene with 100 images from the Blender dataset.
\textbf{Left}: When provided noisy observations of the true camera locations, BARF~\cite{lin2021barf} cannot converge to the correct poses.
\textbf{Middle}: GNeRF~\cite{meng2021gnerf} assumes a 2D camera representation (azimuth, elevation) which is accurate for the Blender dataset which has
that exact configuration (upright cameras on a sphere). However, GNeRF also requires an accurate prior distribution on poses
for sampling. The Lego images live on one hemisphere, but when GNeRF's prior distribution is the full sphere it also fails
to localize the images accurately.
\textbf{Right}: Our full model, LU-NeRF+Sync, is able to recover poses almost perfectly in this particular example. By taking a local-to-global approach, we avoid having
strong assumptions about camera representation or pose priors.
Following~\cite{lin2021barf,meng2021gnerf} pose errors for each method are reported after optimal global alignment of estimated poses to ground truth poses. To
put the translation errors in context, the Blender cameras are on a sphere of radius $4.03$.}
\label{fig:teaser}
\end{figure*}

\section{Introduction}
\label{sec:intro}
NeRF~\cite{mildenhall2020nerf} was introduced as a powerful method to tackle the problem of learning neural scene representations and photorealistic view synthesis, and subsequent research has focused on addressing its limitations to extend its applicability to a wider range of use cases (see~\cite{advances_in_neural_rendering,neuralfieldssurvey} for surveys).
One of the few remaining hurdles for view synthesis in the wild is the need for accurate localization.  
As images captured in the wild have unknown poses, these approaches often use structure-from-motion (SfM)~\cite{schoenberger2016sfm,ozyesil2017sfmsurvey} to determine the camera poses.
There is often no recourse when SfM fails (see Fig.~\ref{fig:colmap} for an example), 
and in fact, even small inaccuracies in camera pose estimation can have a dramatic impact on photorealism.

Few prior attempts have been made to reduce the reliance on SfM by integrating pose estimation directly within the NeRF framework. 
However, the problem is severely underconstrained (see Fig.~\ref{fig:teaser}) and current approaches make additional assumptions to make the problem tractable. For example, NeRf$--$~\cite{wang2021nerfmm} focuses on pose estimation in forward-facing configurations, BARF~\cite{lin2021barf} initialization must be close to the true poses, and GNeRF~\cite{meng2021gnerf} assumes a 2D camera model (upright cameras on a hemisphere).

We propose an approach for jointly estimating the camera pose and scene representation from images from a single scene while allowing for a more general camera configuration than previously possible.
Conceptually, our approach is organized in a local-to-global learning framework using NeRFs.
In the \emph{local} processing stage we partition the scene into overlapping subsets, each containing only a few images (we call these subsets \emph{mini-scenes}). Knowing images in a mini-scene are mostly nearby is what makes the joint estimation of pose and scene better conditioned than performing the same task globally.
In the \emph{global} stage, the overlapping mini-scenes are registered in a common reference frame through pose synchronization, followed by jointly refining all poses and learning the global scene representation.

This organization into mini-scenes requires learning from a few local unposed images. Although methods exist for few-shot novel view synthesis~\cite{yu2020pixelnerf,kulhanek2022viewformer,Niemeyer2021Regnerf,kangle2021dsnerf,chen2022geoaug,chen2021mvsnerf}, and separately for optimizing unknown poses~\cite{lin2021barf,meng2021gnerf,wang2021nerfmm}, the combined setting presents new challenges.
Our model must reconcile the ambiguities prevalent in the local unposed setting -- in particular the mirror symmetry ambiguity~\cite{Ozden2010MultibodySI}, where two distinct 3D scenes and camera configurations produce similar images under affine projection. 

We introduce a Local Unposed NeRF (LU-NeRF) model to address these challenges in a principled way.  The information from the LU-NeRFs (estimated poses, confidences, and mirror symmetry analysis) is used to register all cameras in a common reference frame through pose synchronization~\cite{dellaert2020shonan,se3sync,hartley13ijcv}, after which we refine the poses and optimize the neural scene representations using all images.
In summary, our key contributions are:
\begin{itemize}[itemsep=1pt,topsep=0pt]
\item A local-to-global pipeline that learns both the camera poses in a general configuration and a neural scene representation from only an unposed image set.
\item LU-NeRF, a novel model for few-shot local unposed NeRF. LU-NeRF is tailored to the unique challenges we have identified in this setting, such as reconciling mirror-symmetric configurations.
\end{itemize}
Each phase along our local-to-global process is designed with robustness in mind, and the consequence is that our pipeline can be successful even when the initial mini-scenes contain frequent outliers (see Sec~\ref{sec:exp} for a discussion on different mini-scene construction techniques).  The performance of our method surpasses prior works that jointly optimize camera poses and scene representation, while also being flexible enough to operate in the general SE(3) pose setting unlike prior techniques. Our experiments indicate that our pipeline is complementary to the feature-based SfM pipelines used to initialize NeRF models, and is more reliable in low-texture or low-resolution settings.

\section{Related work}
\label{sec:related}
\noindent
\textbf{Structure from motion (SfM).} 
Jointly recovering 3D scenes and estimating camera poses from multiple views of a scene is the classic problem in Computer Vision~\cite{mvgboook}.  Numerous techniques have been proposed for SfM~\cite{ozyesil2017sfmsurvey,schoenberger2016sfm} with unordered image collections and visual-SLAM for sequential data~\cite{taketomi2017visual,murTRO2015}. These techniques are largely built upon local features~\cite{lowe2004sift,rublee2011orb,detone2018superpoint,sun2021loftr} and require accurate detection and matching across images.  
The success of these techniques has led to their widespread adoption, and existing deep-learning approaches for scene representation and novel view synthesis are designed with the implicit assumption that the SfM techniques provide accurate poses in the wild. For example, NeRF~\cite{mildenhall2020nerf} and its many successors (\eg \cite{barron2021mip,barron2022mipnerf360,mueller2022instantngp}) utilize poses estimated offline with COLMAP~\cite{schoenberger2016sfm,lindenberger2021pixel}. 
However, COLMAP can fail on textureless regions and low-resolution images. 

The local-to-global framework proposed in this work is inspired by the ``divide-and-conquer'' SfM and SLAM methods~\cite{bhowmick2015divide,zhu2018very,fang2019merge,chen2020graph,cucuringu2012sensor,zhou2020stochastic,cin2021synchronization}.

\begin{figure*}[t]
\centering
\includegraphics[width=\linewidth]{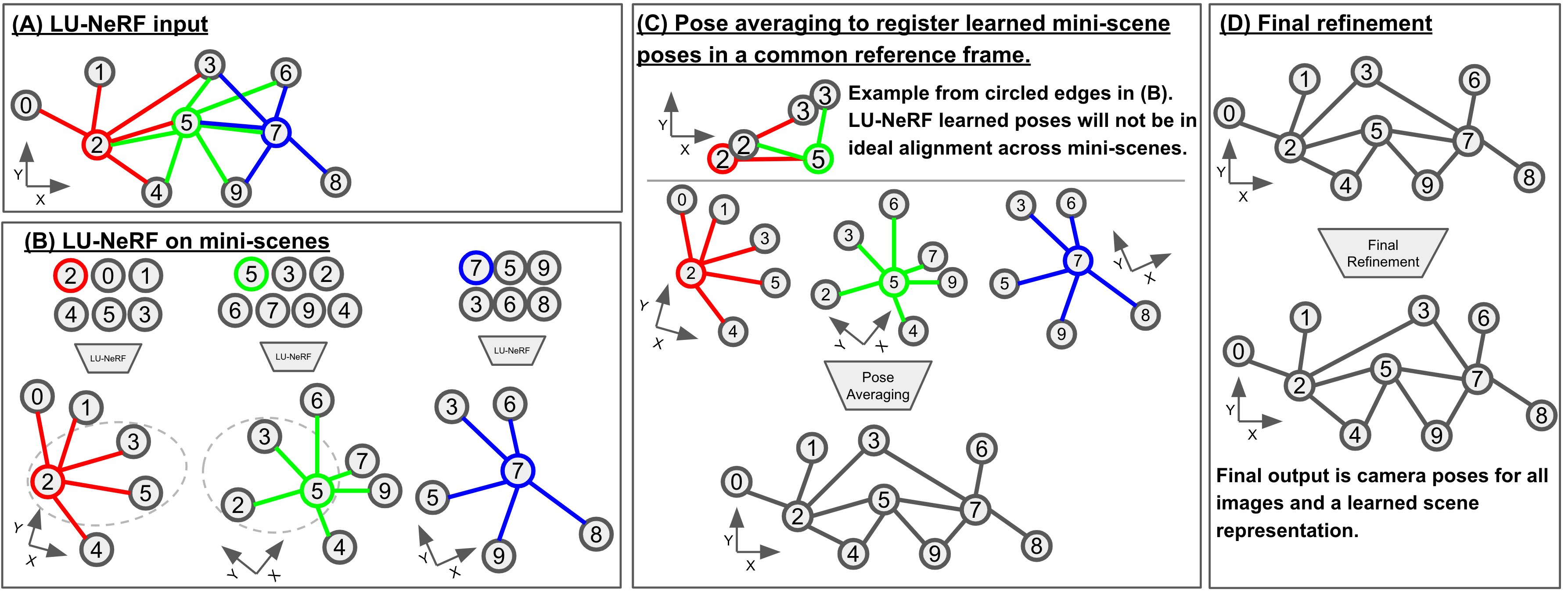}
\caption{\textbf{Proposed method.} (A) shows the ground truth locations of each image (we show this only for visualization). Edge colors show the grouping within mini-scenes. We create a mini-scene for each image, though here only three mini-scenes are highlighted; the ones centered at image 2 (red edges), image 5 (green edges), and image 7 (blue edges). Depending on the strategy used to create mini-scenes, the grouped images can contain outlier images far from the others.
(B) LU-NeRF takes unposed images from a single mini-scene and optimizes poses without any constraints on the pose representation.
(C) The reference frame and scene scale learned by LU-NeRF is unique to each mini-scene. This, plus estimation errors, means the relative poses between images in overlapping mini-scenes will not perfectly agree.  To register the cameras in a common reference frame, we utilize pose synchronization which seeks a globally optimal positioning of all cameras from noisy relative pose measurements -- this is possible since we have multiple relative pose estimations for many pairs of images. 
(D) Lastly, we jointly refine the synchronized camera poses and learn a scene representation.
}
\label{fig:overview}
\end{figure*}

\noindent
\textbf{Neural scene representation with unknown poses.}
BARF~\cite{lin2021barf} and GARF~\cite{chng2022garf} jointly optimize neural scene and camera
poses, but require good initialization (\eg within \ang{15} of the groundtruth). 
NeRF$--$~\cite{wang2021nerfmm}, X-NeRF~\cite{poggi2022cross}, SiNeRF~\cite{xia2022sinerf}, and SaNeRF~\cite{Pitteri2019} only work on forward-facing scenes; 
SAMURAI~\cite{samurai2022} aims to handle coarsely specified poses (octant on a sphere) using a pose multiplexing strategy during training;
GNeRF~\cite{meng2021gnerf} and VMRF~\cite{vmrf} are closest to our problem setting.
They do not require accurate initialization and work on \ang{360} scenes. However, they make strong assumptions about the pose distribution, assuming 2DoF and a limited elevation range.
Performance degrades when the constraints are relaxed. 

Approaches that combine visual SLAM with neural scene representations~\cite{zhu2022nice,sucar2021imap,rosinol2022nerf} typically rely on RGB-D streams and are exclusively designed for video sequences. The use of depth data significantly simplifies both scene and pose estimation processes.  There are several parallel efforts to ours in this field. For instance, NoPe-NeRF~\cite{bian2022nope} trains a NeRF without depending on pose priors; however, it relies on monocular depth priors. In a manner akin to our approach, LocalRF~\cite{meuleman2023progressively} progressively refines camera poses and radiance fields within local scenes. Despite this similarity, it presumes monocular depth and optical flow as supervision, and its application is limited to ordered image collections; 
MELON~\cite{levy2023melon} optimizes NeRF with unposed images using equivalence class estimation, yet it is limited to \SO(3); RUST~\cite{sajjadi2022rust} and FlowCam~\cite{smith2023flowcam} learn a generalizable neural scene representation from unposed videos.

In summary, prior work on neural scene representation with unknown poses assumes either small
perturbations~\cite{lin2021barf,chng2022garf,wang2021nerfmm,xia2022sinerf}, a narrow distribution of camera poses~\cite{meng2021gnerf,vmrf}, or depth priors~\cite{bian2022nope,meuleman2023progressively}.
To the best of our knowledge, we are the first to address the problem of neural rendering with
unconstrained unknown poses for both ordered and unordered image collections.

\noindent
\textbf{Few-shot scene estimation.}
Learning scene representations from a few images has been studied
in~\cite{yu2020pixelnerf,kangle2021dsnerf,chen2022geoaug,chen2021mvsnerf,kulhanek2022viewformer,Niemeyer2021Regnerf}.
PixelNeRF~\cite{yu2020pixelnerf} uses deep CNN features to construct NeRFs from few or even a single image.
MVSNeRF~\cite{chen2021mvsnerf} leverages cost-volumes typically applied in multi-view stereo for
the same task, while DS-NeRF~\cite{kangle2021dsnerf} assumes depth supervision is available to enable
training with fewer views.
Our approach to handle the few-shot case relies on a standard neural field optimization
with strong regularization, similar to RegNeRF~\cite{Niemeyer2021Regnerf}.

\noindent
\textbf{Unsupervised pose estimation.} 
There are a number of techniques that can learn to predict object pose from categorized image collections without explicit pose supervision. Multiple views of the same object instance are used in~\cite{mvcTulsiani18,insafutdinov18neurips} to predict the shape and pose while training is self-supervised through shape rendering. 
RotationNet~\cite{kanezaki2018cvpr} uses multiple views of an object instance to predict both poses and class labels but is limited to a small set of discrete uniformly spaced camera viewpoints.
The multi-view input is relaxed in ~\cite{monnier2022unicorn,wu2020unsup3d} which operates on single image collections for a single category. UNICORN~\cite{monnier2022unicorn} learns a disentangled representation that includes pose and utilizes cross-instance consistency at training, while an assumption about object symmetry guides the training in~\cite{wu2020unsup3d}.

\section{Methodology}
\label{sec:method}

An illustration of our approach is shown in Figure~\ref{fig:overview}. At the core of our method is the idea of breaking up a large scene into mini-scenes
to overcome the non-convexity of global pose optimization without accurate initialization.
When the camera poses in the mini-scene are close to one another, we are able to initialize the
optimization with all poses close to the identity and optimize for relative poses. In Sec.~\ref{sec:exp}, we describe how we construct mini-scenes, and below we describe the process of local shape estimation followed by global synchronization.

\subsection{Local pose estimation}
The local pose estimation step takes in mini-scenes of typically three to five images
and returns the relative poses between the images.
The model, denoted LU-NeRF-1,
is a small NeRF~\cite{mildenhall2020nerf} that jointly optimizes the camera poses as extra
parameters as in BARF~\cite{lin2021barf}. 
In contrast with BARF, in this stage, we are only interested in a rough pose estimation that
will be improved upon later, so we aim for a lightweight model with faster convergence by using
small MLPs and eliminating positional encoding and view dependency.
As we only need to recover relative poses, without loss of generality, we freeze one of the poses at identity and optimize all the others.

Few-shot radiance field optimization is notoriously difficult and requires strong regularization~\cite{Niemeyer2021Regnerf}.
Besides the photometric $\ell_2$ loss proposed in NeRF, 
we found that adding a loss term for the total variation of the predicted depths over small patches
is crucial for the convergence of both camera pose and scene representation:
\small
\begin{equation*}
    \frac{1}{|\mathcal{R}|}\sum_{\mathbf{r} \in \mathcal{R}} \sum_{i,j=1}^{K} \big(d_\theta(\mathbf{r}_{i,j}) - d_\theta(\mathbf{r}_{i,j+1})\big)^2 + \big(d_\theta(\mathbf{r}_{i,j}) - d_\theta(\mathbf{r}_{i+1,j})\big)^2
\end{equation*}
\normalsize
where $\mathcal{R}$ is a set of ray samples, $d_\theta(\mathbf{r})$ is the depth rendering function for a ray $\mathbf{r}$, $\theta$ are the model parameters and camera poses, $K$ is the patch size, and $(i,j)$ is the pixel index. 

\subsection{Mirror-symmetry ambiguity}\label{sec:mirror-symmtry}
\begin{figure}[h]
\centering
\includegraphics[width=\columnwidth]{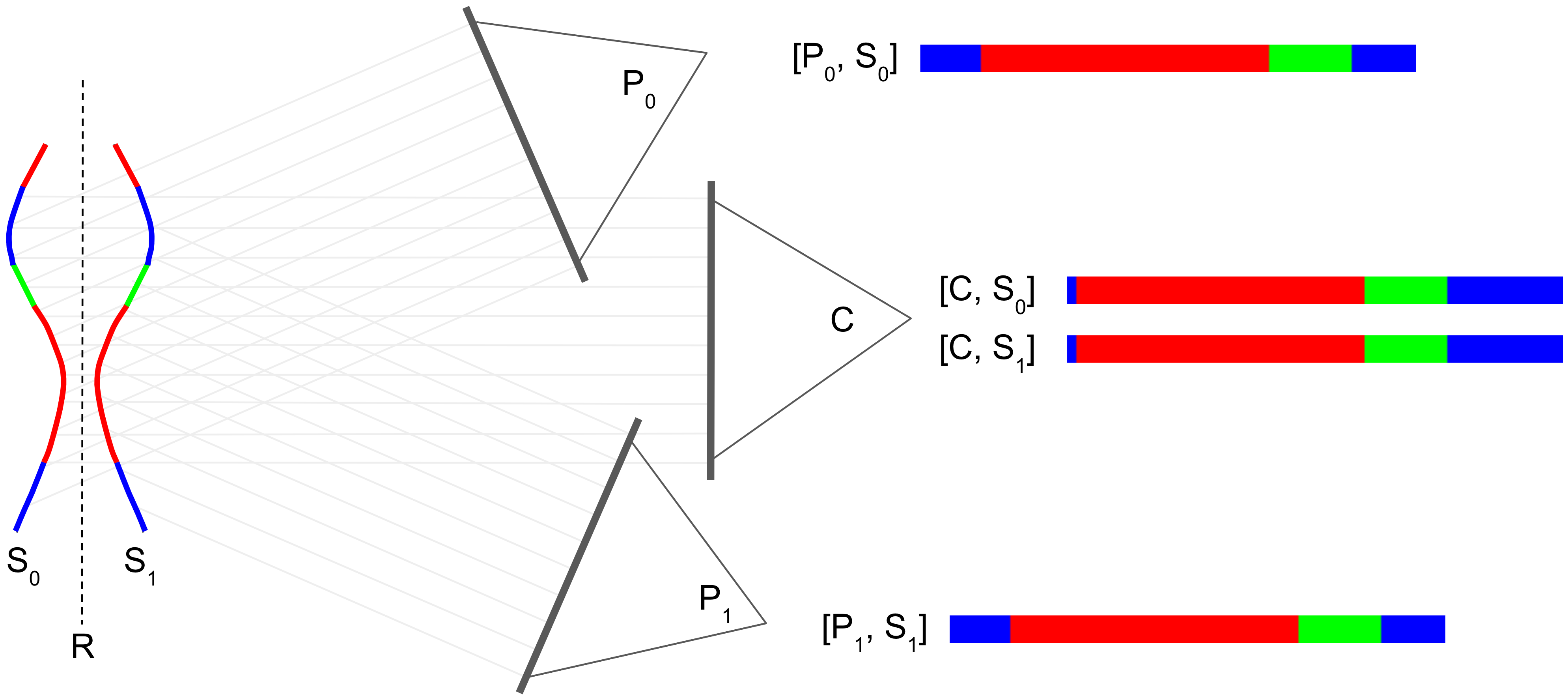}
\caption{\textbf{Mirror symmetry ambiguity.}
Under affine projection, a 3D scene ($S_0$) and its reflection ($S_1$) across a plane ($R$) will produce the same image viewed from affine camera $C$.
The consequence of this is that two distinct 3D scenes and camera poses will produce similar images. In this illustration, scene $S_0$ viewed from camera $P_0$ will produce the same image as the reflected scene $S_1$ viewed from $P_1$.  While this relationship is exact in the affine model, we observe that the mini-scene configuration with respect to the scene structure is often well-approximated as affine and training can converge to the near-symmetric solutions. Our LU-NeRF model is explicitly designed to anticipate this failure mode. This illustration is inspired by a similar diagram in~\cite{Ozden2010MultibodySI}.}
\label{fig:mirror}
\end{figure}

The ambiguities and degeneracies encountered when estimating 3D structure have been extensively studied~\cite{szeliski1997shape,belhumeur1999bas,chum-degen-cvpr05}. One particularly relevant failure mode of SfM is distant small objects, where the perspective
effects are small and can be approximated by an affine transform, and one cannot differentiate
between reflections of the object around planes parallel to the image plane~\cite{Ozden2010MultibodySI}.
When enforcing multi-view consistency, this effect, known as mirror-symmetry ambiguity,
can result in two different configurations of structure and motion that cannot be told apart
(see \cref{fig:mirror}). We notice, perhaps for the first time, that neural radiance fields with unknown poses can degenerate in the same way.

One potential solution to this problem would be to keep the two possible solutions
and drop one of them when new observations arrive.
This is not applicable to our case since at this stage the only information available is the
few images of the mini-scene.

To mitigate the issue, we introduce a second stage for the training, denoted LU-NeRF-2.
We take the estimated poses in world-to-camera frame $\{R_i\}$ from LU-NeRF-1,
and the reflected cameras $\{R_{\pi} R_i\}$, where $R_\pi$ is a rotation around the optical axis.
Note that this is different than post-multiplying by $R_\pi$, which would correspond to a
global rotation that wouldn't change the relative poses that we are interested in at this stage.
We then train two new models, with the scene representation started from scratch and poses initialized as the original and reflected sets,
and resolve the ambiguity by picking the one with the smallest photometric training loss.
The rationale is that while the issue is caused by LU-NeRF-1 ignoring small perspective distortions,
the distortions can be captured on the second round of training, which is easier since one of the initial sets of poses is expected to be reasonable.

\subsection{Local to global pose estimation}
After training LU-NeRF-2, we have sets of relative poses for each mini-scene in some local frame.
The problem of finding a global alignment given a set of noisy relative poses is known as
pose synchronization or pose averaging.
It is formalized as optimizing the set of $N$ global poses $\{P_i\}$ given relative pose observations
$R_{ij}$,
\begin{align}
  \argmin_{P \in \SE(3)^N} d(P_{ij}, P_jP_i^\top),
\end{align}
for some metric $\fun{d}{\SE(3) \times \SE(3)}{\R}$.
The problem is challenging due to non-convexity and
is an active subject of research~\cite{spectral_sync,se3sync,dellaert2020shonan}.
We use the Shonan rotation method~\cite{dellaert2020shonan} to estimate the camera rotations, followed by a least-squares optimization of the translations.

\noindent
\textbf{Global pose and scene refinement.}
After pose averaging, the global pose estimates are expected to be good enough such
that any method that requires cameras initialized close to the ground truth should work (\eg BARF~\cite{lin2021barf}, GARF~\cite{chng2022garf}). 
We apply BARF~\cite{lin2021barf} at this step, which results in both accurate poses
and a scene representation accurate enough for realistic novel view synthesis.
We refer to the full pipeline as LU-NeRF+Sync.

\section{Experiments}
\label{sec:exp}

\begin{table*}[h]
\resizebox{\textwidth}{!}{
\centering
\begin{tabular}{@{}
  l
  ccr
  ccr
  ccr
  ccr
  ccr
  cc
  @{}}
\toprule
  & \multicolumn{2}{c}{Chair} &               & \multicolumn{2}{c}{Hotdog} &               & \multicolumn{2}{c}{Lego} &  & \multicolumn{2}{c}{Mic} &               & \multicolumn{2}{c}{Drums} &               & \multicolumn{2}{c}{Ship}                                            \\
  \cmidrule{2-3} \cmidrule{5-6} \cmidrule{8-9} \cmidrule{11-12} \cmidrule{14-15} \cmidrule{17-18}

                  & rot                       & trans         &                            & rot           & trans                    &  & rot                     & trans         &                           & rot           & trans         &  & rot           & trans         &  & rot   & trans \\
\hline
COLMAP            & 0.12                      & 0.01          &                            & 1.24          & 0.04                     &  & 2.29                    & 0.10          &                           & 8.37          & 0.18          &  & 5.91          & 0.28          &  & 0.17  & 0.01  \\
\quad +BARF         & 0.14                      & 0.01          &                            & 1.20          & 0.01                     &  & 1.88                    & 0.09          &                           & 3.73          & 0.15          &  & 8.71          & 0.54          &  & 0.15  & 0.01  \\
  \hline
VMRF \ang{120}    & 4.85                      & 0.28          &                            & --             & --                        &  & 2.16                    & 0.16          &                           & 1.39          & 0.07          &  & 1.28          & 0.08          &  & 16.89 & 0.71  \\
GNeRF \ang{90}    & 0.36                      & 0.02          &                            & 2.35          & 0.12                     &  & 0.43                    & 0.02          &                           & 1.87          & 0.03          &  & 0.20          & 0.01          &  & 3.72  & 0.18  \\
GNeRF \ang{120}   & 4.60                      & 0.16          &                            & 17.19         & 0.74                     &  & 4.00                    & 0.20          &                           & 2.44          & 0.08          &  & 2.51          & 0.11          &  & 31.56 & 1.38  \\
GNeRF \ang{150} & 16.10                     & 0.76          &                            & 23.53         & 0.92                     &  & 4.17                    & 0.36          &                           & 3.65          & 0.26          &  & 5.01          & 0.18          &  & --     & --     \\
\hline
GNeRF \ang{180} (2DOF)   & 24.46                     & 1.22          &                            & 36.74         & 1.46                     &  & 8.77                    & 0.53          &                           & 12.96         & 0.66          &  & \textbf{9.01} & 0.49          &  & --     & --     \\
Ours (3DOF)              & \textbf{2.64}             & \textbf{0.09} &                            & \textbf{0.24} & \textbf{0.01}            &  & \textbf{0.09}           & \textbf{0.00} &                           & \textbf{6.68} & \textbf{0.10} &  & 12.39         & \textbf{0.23} &  & --     & --     \\
\bottomrule
\end{tabular}
}
\caption{\textbf{Camera pose estimation on unordered image collection.}
GNeRF~\cite{meng2021gnerf} and VMRF~\cite{vmrf} constrain the elevation range,
where the maximum elevation is always \ang{90}.
For example, GNeRF \ang{120} only samples elevations in [\ang{-30,}, \ang{90}].
The \ang{180} variations don't constrain elevation and are closest to our method,
but they are still limited to 2 degrees of freedom for assuming upright cameras.
Bold numbers indicate superior performance between the bottom two rows, which are the fairest comparison among NeRF-based methods, although our method is still solving a harder 3DOF problem
  versus 2DOF of GNeRF.
We outperform GNeRF in all but one scene in this comparison.
  COLMAP~\cite{schoenberger2016sfm} results in its best possible scenario are shown for reference (higher resolution images and assuming
  optimal graph to set unregistered poses to the closest registered pose). COLMAP+BARF runs a BARF refinement
  on top of these initial results, and even in this best-case scenario, our method still outperforms
  it in some scenes, which shows that LU-NeRF can complement COLMAP and work in scenes COLMAP fails.
  Our model fails on the Ship scene due to outliers in the connected graph; GNeRF with fewer constraints
  also fails on it.
  We provide a detailed error analysis on the Drums scene in the Appendix. 
}
\label{tab:unorder-pose}
\end{table*}

\begin{figure*}[h]
\centering
\includegraphics[width=\linewidth]{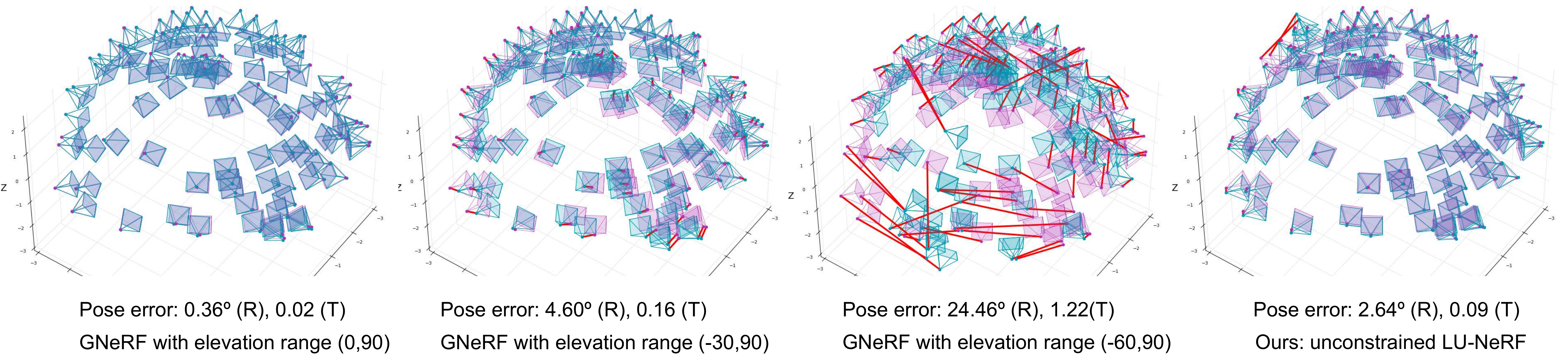}
\caption{\textbf{Camera pose estimation on unordered image collections.} The performance of GNeRF drops dramatically when the pose prior is expanded beyond the true distribution. In comparison, our method does not rely on any prior knowledge of pose distribution.}
\label{fig:camera-unorder}
\end{figure*}

\begin{table*}[t]
\resizebox{\textwidth}{!}{
  \centering
\begin{tabular}{@{}lcccccccccccccccc@{}}
  \toprule
                       & \multicolumn{3}{c}{Chair} &          & \multicolumn{3}{c}{Drums} &  & \multicolumn{3}{c}{Lego} &          & \multicolumn{3}{c}{Mic}                                                             \\
 \cmidrule{2-4}\cmidrule{6-8}\cmidrule{10-12}\cmidrule{14-16}
                       & PSNR $\uparrow$                  & SSIM $\uparrow$ & LPIPS $\downarrow$                 &  & PSNR $\uparrow$                 & SSIM $\uparrow$ & LPIPS $\downarrow$ &  & PSNR $\uparrow$ & SSIM $\uparrow$ & LPIPS $\downarrow$ &  & PSNR $\uparrow$ & SSIM $\uparrow$ & LPIPS $\downarrow$ \\
  \midrule
GNeRF \ang{90}         & 31.30                     & 0.95     & 0.08                      &  & 24.30                    & 0.90     & 0.13      &  & 28.52    & 0.91     & 0.09      &  & 31.07    & 0.96     & 0.06      \\
GNeRF \ang{120}        & 25.01                     & 0.89     & 0.15                      &  & 20.63                    & 0.86     & 0.20      &  & 22.95    & 0.85     & 0.16      &  & 23.68    & 0.93     & 0.11      \\
GNeRF \ang{150}        & 22.18                     & 0.88     & 0.20                      &  & 19.05                    & 0.83     & 0.27      &  & 21.39    & 0.84     & 0.18      &  & 23.22    & 0.92     & 0.13      \\
VMRF \ang{120}         & 26.05                     & 0.90     & 0.14                      &  & 23.07                    & 0.89     & 0.16      &  & 25.23    & 0.89     & 0.12      &  & 27.63    & 0.95     & 0.08      \\
VMRF \ang{150}         & 24.53                     & 0.90     & 0.17                      &  & 21.25                    & 0.87     & 0.21      &  & 23.51    & 0.86     & 0.14      &  & 24.39    & 0.94     & 0.10      \\
\midrule
GNeRF \ang{180} (2DOF) & 21.27                     & 0.87     & 0.23                      &  & 18.08                    & 0.81     & 0.33      &  & 18.22    & 0.82     & 0.24      &  & 17.22    & 0.86     & 0.32      \\
VMRF \ang{180} (2DOF)  & 23.18                     & 0.89     & 0.16                      &  & 20.01                    & 0.84     & 0.29      &  & 21.59    & 0.83     & 0.18      &  & 20.29    & 0.90     & 0.22      \\

Ours (3DOF)  & \textbf{30.57} & \textbf{0.95} & \textbf{0.05} && \textbf{23.53} &          \textbf{0.89} &  \textbf{0.12} && \textbf{28.29} & \textbf{0.92} & \textbf{0.06} && \textbf{22.58} & \textbf{0.91} & \textbf{0.08} \\
\bottomrule
\end{tabular}
}
\caption{\textbf{Novel view synthesis on unordered collections.}
Our method outperforms the baselines on most scenes while being more general for considering
arbitrary rotations with 3 degrees-of-freedom.
Here we quote the baseline results from VMRF~\cite{vmrf},
where \emph{hotdog} is not available. We provided the results on all scenes (including \emph{hotdog}) using the public source code of GNeRF in the Appendix. 
}
\label{tab:unorder-view}
\end{table*}

\begin{figure}[t]
\centering
\includegraphics[width=0.9\linewidth]{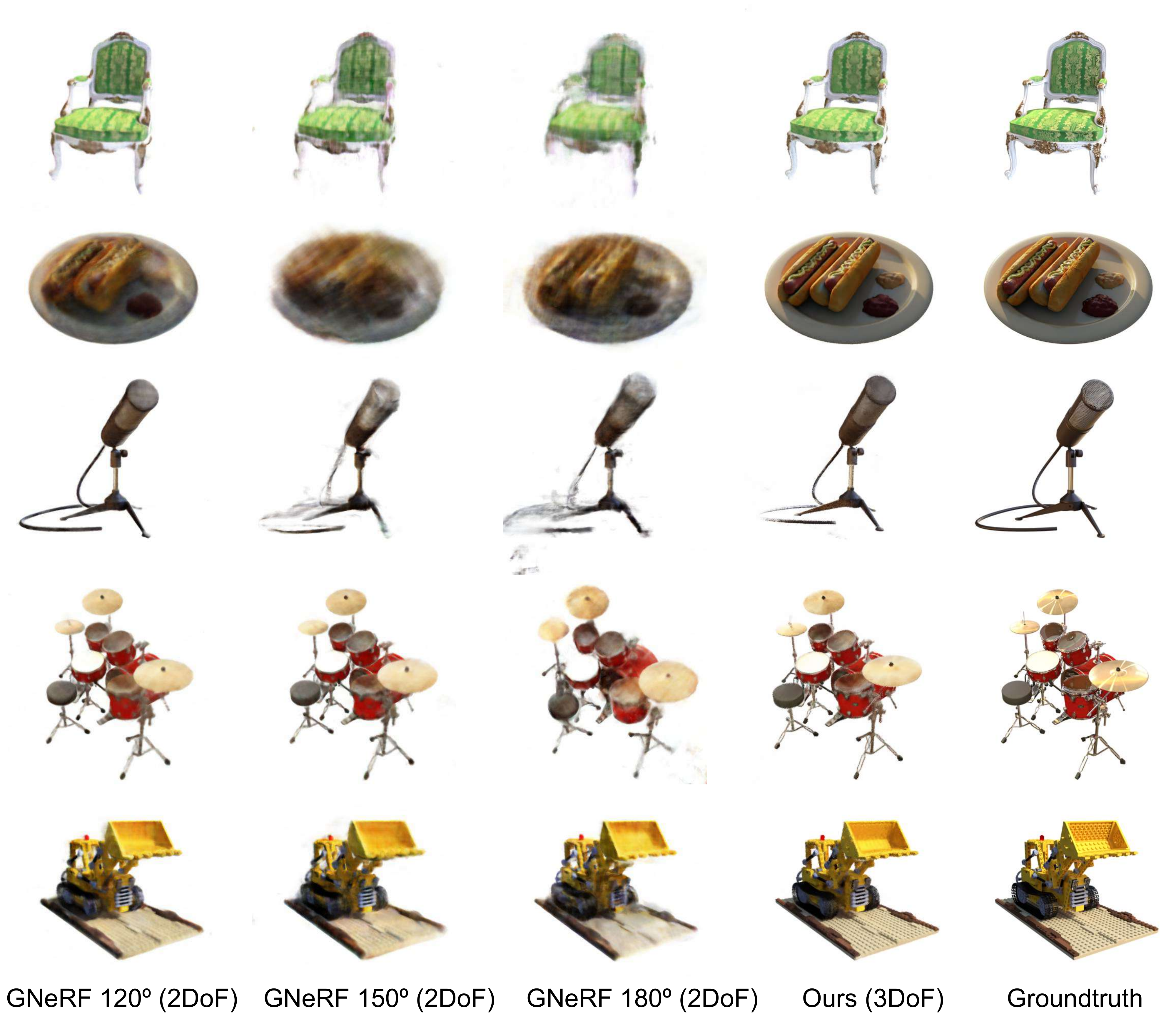}
\caption{\textbf{Novel view synthesis on unordered image collections}. 
GNeRF makes assumptions on the elevation range, where the maximum elevation
is always \ang{90}.
For instance, GNeRF~\ang{150} only samples elevations in [-\ang{60}, \ang{90}].
The \ang{180} variations don’t constrain elevation and are closest to our method, but they are still limited to 2 degrees of freedom for assuming upright cameras.
The performance of GNeRF drops as prior poses are less constrained. 
Please zoom into the figure to see the details in the renderings.}
\label{fig:novel-view-vis}
\end{figure}

\begin{table}[t]
\setlength{\tabcolsep}{3pt}
\centering
  \begin{tabular}{c c c c c c c}
  \toprule
  Image size & Chair & Hotdog & Lego & Mic & Drums & Ship \\
  \midrule
  400$\times$400 & 100 & 88 & 100 & 15 & 74 & 45 \\
  800$\times$800 & 100 & 98 & 100 & 80 & 84 & 100 \\
  \bottomrule
  \end{tabular}
  \caption{\textbf{Number of images registered by COLMAP on Blender.} \label{tab:num_colmap}}
\end{table}

Our method as described in~\cref{sec:method} starts from a set of mini-scenes
that covers the input scene.
We evaluate different approaches to constructing mini-scenes, each with different assumptions on the input.

The most strict assumption is that we have an \emph{optimal graph} connecting each image to
its nearest neighbors in camera pose space.
While this seems unfeasible in practice, some real-life settings approximate this,
for example, when images are deliberately captured in a pattern such as a grid,
or if they are captured with camera arrays.

In a less constrained version of the problem, we assume
an \textit{ordered image collection}, where the images form a sequence,
from where a line graph is trivially built.
This is a mild assumption that is satisfied by video data,
as well as the common setting of a camera physically moving around a scene
sequentially capturing images.

In the most challenging setting,
we assume nothing about the scene and only take an \textit{unordered image collection}.

\noindent
\textbf{Building graphs from unordered image collections.}
We evaluate two simple ways of building graphs from unordered image collections.
The first is to use deep features from a self-supervised model trained on large image collections.
We use the off-the-shelf DINO model~\cite{caron2021emerging,amir2021deep} to extract image features and build the graph based on the cosine distance between these features.
The second is to simply use the $\ell_1$ distance in pixel space against slightly shifted and rotated versions of the images.
Neither of these approaches is ideal. The deep features are typically coarse
and too general, failing to detect specific subtle changes on the scene.
The $\ell_1$ distance has the opposite issue, where small changes can result in large distances.
We provide a detailed analysis in the Appendix.
Exploring other methods for finding a proxy metric for the relative pose in image space
is a direction for future work. 

\noindent
\textbf{Datasets.}
We compare with existing published results on the synthetic-NeRF dataset~\cite{mildenhall2020nerf}. 
We use the training split of the original dataset as our \textit{unordered image collection} which consists of 100 unordered images per 3D scene. 
We use the first 8 images from the validation set as our test set for the novel view synthesis task, following prior works~\cite{meng2021gnerf,vmrf}

To evaluate on image sequences, where the order of images is known,
we further render a Blender \textit{ordered image collection} with 100 images along
a spiral path per scene.
The images are resized to $400\times 400$ in our experiments. 

We also evaluate on real images from the object-centric videos in Objectron~\cite{ahmadyan2021objectron}. The dataset provides ground truth poses computed using AR solutions at 30fps, and we construct a wider-baseline dataset by subsampling every 15th frame and selecting videos with limited texture (Fig.~\ref{fig:colmap}).

\noindent
\textbf{Evaluation metrics.}
We evaluate the tasks of camera pose estimation and novel view synthesis. 
For camera pose estimation, we report the camera rotation and translation error using Procrustes analysis as in BARF~\cite{lin2021barf}.
For novel view synthesis, we report the PSNR, SSIM, and LPIPS~\cite{Zhang_2018_CVPR}.

\noindent
\textbf{Baseline methods.}
We compare with GNeRF~\cite{meng2021gnerf}, VMRF~\cite{vmrf}, and COLMAP~\cite{schoenberger2016sfm} throughout our experiments. 
GNeRF samples camera poses from a predefined prior pose distribution and trains a GAN-based neural rendering model to build the correspondence between the sampled camera poses and 2D renderings. 
The method provides accurate pose estimation under \textit{proper} prior pose distribution. However, its performance degrades significantly when the prior pose distribution doesn't match the groundtruth. 
VMRF attempts to relieve the reliance of GNeRF on the prior pose distribution but still inherits its limitations. 
In our experiments, we evaluate with the default pose priors of GNeRF on the NeRF-synthetic dataset, \ie, azimuth $\in [\ang{0}, \ang{360}]$ and elevation $\in [\ang{0}, \ang{90}]$, and also on less constrained cases.
COLMAP works reliably in texture-rich scenes but may fail dramatically on texture-less surfaces.

\noindent
\textbf{Implementation details.}
We use a compact network for LU-NeRF to speed up the training and minimize the memory cost. 
Specifically, we use a 4-layer MLP without positional encoding and conditioning on the view directions.  
We stop the training early when the change of camera poses on mini-scenes is under a predefined threshold.
To resolve the mirror symmetry ambiguity (Sec.~\ref{sec:mirror-symmtry}), we train two additional LU-NeRFs for a fixed number of training iterations (50k by default). 
The weight of the depth regularization
is 10 times larger than the photometric $\ell_2$ loss throughout our experiments. 
More details are in the Appendix.

\subsection{Unordered Image Collections} 

\begin{table}[t]
\resizebox{\columnwidth}{!}{
\centering
\begin{tabular}{@{}l
  rrl
  rrl
  rrl
  rrl
  rr@{}}
\toprule
  & \multicolumn{2}{c}{Chair} &       & \multicolumn{2}{c}{Drums} &      & \multicolumn{2}{c}{Lego} &  & \multicolumn{2}{c}{Materials} &       & \multicolumn{2}{c}{Mean}          \\
  \cmidrule{2-3}\cmidrule{5-6}\cmidrule{8-9}\cmidrule{11-12}\cmidrule{14-15}
                & rot                       & trans &                           & rot  & trans                    &  & rot                           & trans &  & rot  & trans &  & rot  & trans \\
\midrule
GNeRF \ang{90}  & 11.6                      & 0.49  &                           & 8.03 & 0.29                     &  & 7.89                          & 0.19  &  & 6.80 & 0.12  &  & 8.91 & 0.30  \\
GNeRF \ang{180} & 27.7                      & 1.17  &                           & 130  & 6.23                     &  & 123                           & 4.31  &  & 30.9 & 1.40  &  & 94.9 & 3.27  \\
Ours (3DOF)     & \textbf{0.72}                      & \textbf{0.03}  &                           & \textbf{0.07} & \textbf{0.08}                     &  & \textbf{1.96}                          & \textbf{0.00} &  & \textbf{0.31} & \textbf{0.00}  &  & \textbf{0.76} & \textbf{0.03}  \\
\bottomrule
\end{tabular}
}
\caption{\textbf{Pose estimation on the Blender \emph{ordered image collections}.}
  We report rotation errors in degrees and translation at the input scene scale. 
  Our method can be more easily applied to ordered image collections since the graph-building step becomes trivial.
  In this case, we outperform GNeRF even when it is aided by known and constrained pose distributions.}
\label{tab:ordered-pose}
\end{table}

\begin{table}[!t]
\resizebox{\columnwidth}{!}{
  \centering

\begin{tabular}{@{}lccccccc@{}}
  \toprule
 & \multicolumn{3}{c}{Chair} & & \multicolumn{3}{c}{Drums} \\
  \cmidrule{2-4}\cmidrule{6-8}
  & PSNR $\uparrow$ & SSIM $\uparrow$ & LPIPS $\downarrow$ & & PSNR $\uparrow$ & SSIM $\uparrow$ & LPIPS $\downarrow$ \\
\midrule
GNeRF \ang{90}         &          27.22 &          0.93 &          0.17 &&          20.88 &          0.84 &          0.29 \\
\midrule
GNeRF \ang{180} (2DOF) &          23.50 &          0.91 &          0.26 &&          11.01 &          0.81 &          0.56 \\
Ours (3DOF)            & \textbf{33.94} & \textbf{0.98} & \textbf{0.03} && \textbf{25.29} & \textbf{0.91} & \textbf{0.08} \\
\midrule
& \multicolumn{3}{c}{Lego} & & \multicolumn{3}{c}{Materials} \\
  \cmidrule{2-4}\cmidrule{6-8}
  & PSNR $\uparrow$ & SSIM $\uparrow$ & LPIPS $\downarrow$ & & PSNR $\uparrow$ & SSIM $\uparrow$ & LPIPS $\downarrow$ \\
\midrule
GNeRF \ang{90}  & 22.83 &          0.83 &          0.25 &&          22.58 &          0.85 &          0.20 \\
GNeRF \ang{180} (2DOF) &          9.78 & \textbf{0.78} &          0.53 &&           9.48 &          0.65 &          0.50 \\
Ours (3DOF)            & \textbf{15.90} &          0.72 & \textbf{0.20} && \textbf{29.73} & \textbf{0.96} & \textbf{0.03} \\
\bottomrule
\end{tabular}		
}

\caption{\textbf{Novel view synthesis on Blender \emph{ordered image collections}.}
  The relative improvement of our method with respect to GNeRF is larger with an ordered image collection, since we avoid the difficult step of building the initial graph.
}
\label{tab:view-syntheis-ordered}
\end{table}

\begin{figure}[t]
\centering
\includegraphics[width=\linewidth]{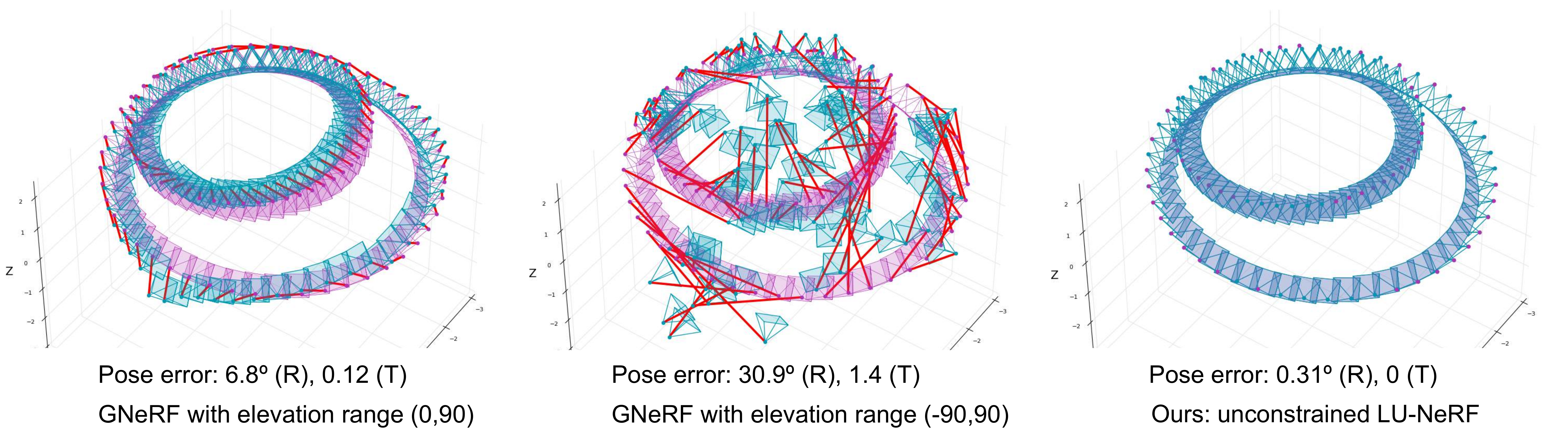}
\caption{\textbf{Pose estimation on the Blender Materials \emph{ordered image collection}.}}
\label{fig:camera-order}
\end{figure}

\noindent
\textbf{Camera pose estimation.}
Tab.~\ref{tab:unorder-pose} compares our method to GNeRF, VMRF, and COLMAP in the camera pose estimation task.
GNeRF achieves high pose estimation accuracy when the elevation angles are uniformly sampled from a \ang{90} interval; however, its performance drops significantly when the range of elevation is enlarged.
Our method outperforms GNeRF in most scenes when the prior pose distribution is unknown,
since we do not require any prior knowledge of the camera poses. 
Fig.~\ref{fig:camera-unorder} provides the visualization of the estimated camera poses from GNeRF under different prior pose distributions and our method. 

Tab.~\ref{tab:num_colmap} shows the number of images COLMAP registers out of 100 in each scene. 
COLMAP is sensitive to image resolution, and its performance drops significantly on low-resolution images. For instance, COLMAP only registers 15 images out of 100 on the Mic scene when the image size is $400\times400$. 
Our method provides accurate pose estimation for all cameras given $400\times400$ images.
Tab.~\ref{tab:unorder-pose} also reports how COLMAP performs in the pose estimation task on the Blender scenes. 
We use the most favorable settings for COLMAP -- $800\times800$ images and set the poses of unregistered cameras to the poses of the nearest registered camera, assuming the \emph{optimal graph} is known, while our method makes no such assumption. Nevertheless, our model achieves better performance than COLMAP in some scenes, even when a BARF refinement is applied to initial COLMAP results.
This shows that LU-NeRF complements COLMAP by working in scenes where COLMAP fails.

\noindent
\textbf{Novel view synthesis.}
Fig.~\ref{fig:novel-view-vis} and Tab.~\ref{tab:unorder-view} show our results
in the task of novel view synthesis on unordered image collections. 
The results are consistent with the quantitative pose evaluation -- our model outperforms both VMRF and GNeRF when no priors on pose distribution are assumed.

\subsection{Ordered Image Collections} 

\subsection{Blender}
Tab.~\ref{tab:ordered-pose}, Tab.~\ref{tab:view-syntheis-ordered}, and Fig.~\ref{fig:camera-order} summarize the results on the Blender \emph{ordered image collection}.  Our method outperforms GNeRF with both constrained
and unconstrained pose distributions even though the elevation of the cameras in this dataset is constrained. Our method utilizes the image order to build a connected graph and does not make any assumptions about the camera distribution.
Results in Tab.~\ref{tab:view-syntheis-ordered} show that the view synthesis results are in sync with the pose estimation results. GNeRF degrades significantly under unconstrained pose priors, while our method outperforms GNeRF consistently across different scenes.

\subsection{Objectron}
\begin{figure}
\centering
\includegraphics[width=\columnwidth]{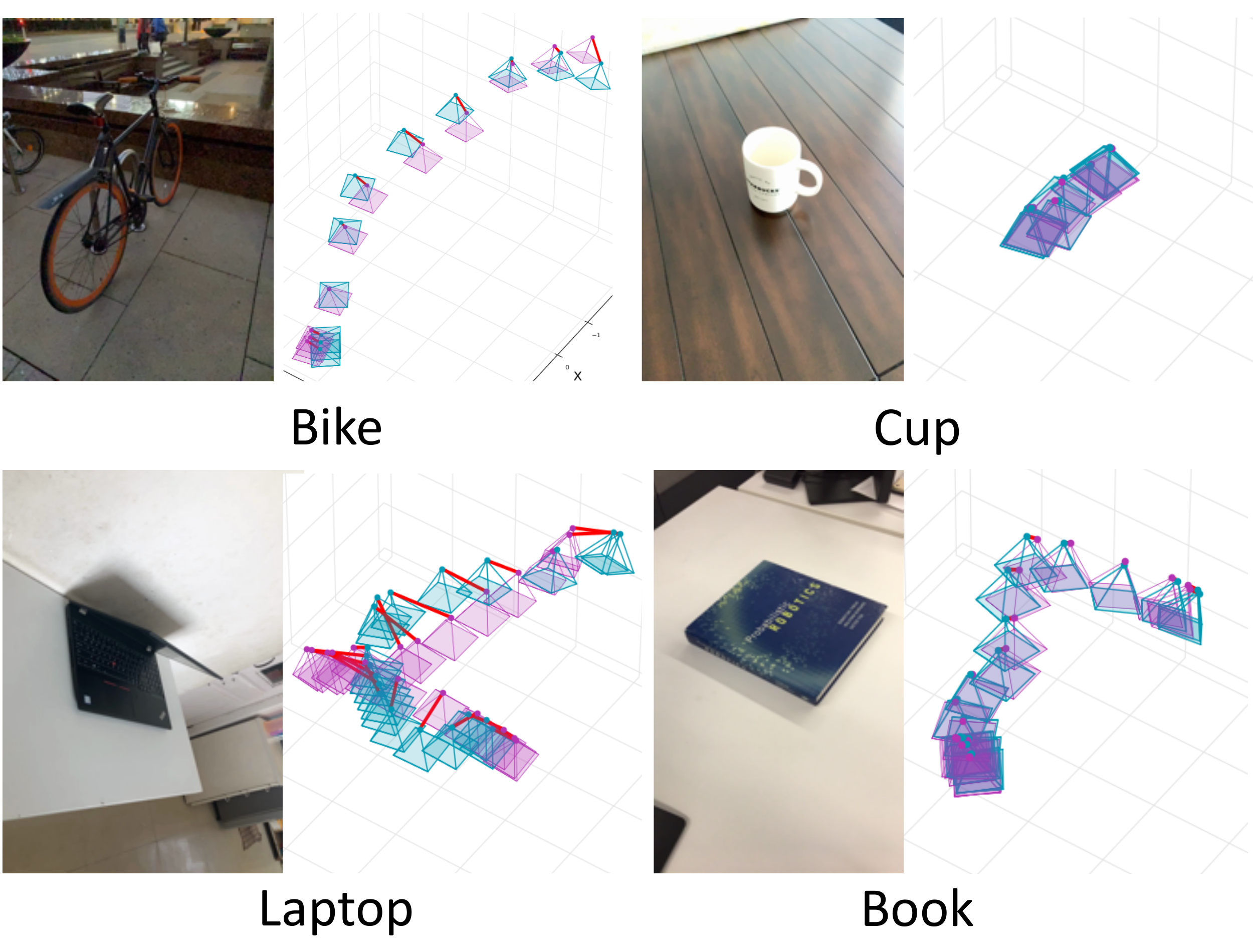}
\caption{\textbf{Camera pose estimation on textureless scenes.}
  COLMAP fails to register any cameras in these Objectron scenes.
  Ground truth cameras are in purple, our predictions in blue.}
\label{fig:colmap}
\end{figure}

We further compare with COLMAP on real images from the Objectron dataset. 
COLMAP can be improved with modern feature extraction and matching algorithms~\cite{sarlin2019coarse} such as SuperPoint~\cite{detone2018superpoint} and SuperGLUE~\cite{sarlin2020superglue} 
(denoted COLMAP-SPSG), or LoFTR~\cite{sun2021loftr} (denoted COLMAP-LoFTR), but these still struggle in scenes with little or repeated texture. Tab.~\ref{tab:real-data} and Fig.~\ref{fig:colmap} show our results \emph{without BARF refinement} on difficult scenes from Objectron.

\begin{table}
\resizebox{\columnwidth}{!}{
\centering
\begin{tabular}{@{}
  l
  cccccc}
\toprule
                    & Bike          & Chair         & Cup           & Laptop        & Shoe          & Book          \\
\midrule
\emph{Rotation:}    &               &               &               &               &               &               \\
\; COLMAP           & --             & 17.2          & --             & --             & 14.1          & --             \\
\; COLMAP-SPSG         & 129           & 28.3          & --             & --             & \textbf{8.3}          & --             \\
\; COLMAP-LoFTR   & \textbf{1.1} & 6.7 & 6.3 & \textbf{9.5} & 14.5 & 83.4  \\
\; Ours             & 15.6 & \textbf{2.6} & \textbf{6.1} & 17.8 & 8.8 & \textbf{3.2} \\
\midrule
\emph{Translation:} &               &               &               &               &               &               \\
\; COLMAP           & --             & 0.04 & --             & --             & \textbf{0.03} & --             \\
\; COLMAP-SPSG         & 1.71          & 0.12          & --             & --             & 0.04          & --             \\
\; COLMAP-LoFTR & \textbf{0.10} & 0.07 & \textbf{0.03} & 0.34 & 0.14 & 0.67    \\
\; Ours             & 0.13 & \textbf{0.03}   & 0.11 & \textbf{0.16} & 0.20          & \textbf{0.03} \\
\bottomrule
\end{tabular}
}
\caption{\textbf{Comparison with COLMAP on Objectron~\cite{ahmadyan2021objectron}.} We report rotation (\textdegree) and translation errors 
  on select scenes from Objectron that are challenging to COLMAP.
  \textbf{``--"} denotes failure to estimate any camera poses. \textbf{COLMAP-SPSG} is an improved version~\cite{sarlin2019coarse} with SuperPoint~\cite{detone2018superpoint} and SuperGLUE~\cite{sarlin2020superglue} as descriptor and matcher, respectively.
  \textbf{COLMAP-LoFTR} improves COLMAP with LoFTR~\cite{sun2021loftr}, a detector-free feature matcher. 
  Translation errors are in the scale of the ground truth scene.
}
\label{tab:real-data}
\end{table}

\subsection{Analysis} 

This section provides additional analysis of our approach. All the experiments discussed below were conducted on the unordered image collection. See the Appendix for an extended discussion.

\begin{figure} [!t]
    \setlength{\tabcolsep}{1mm}
    \newcommand\wifig{0.47\linewidth}
    \centering
    \small
     \begin{tabular}{c|c}
    \includegraphics[width=\wifig]{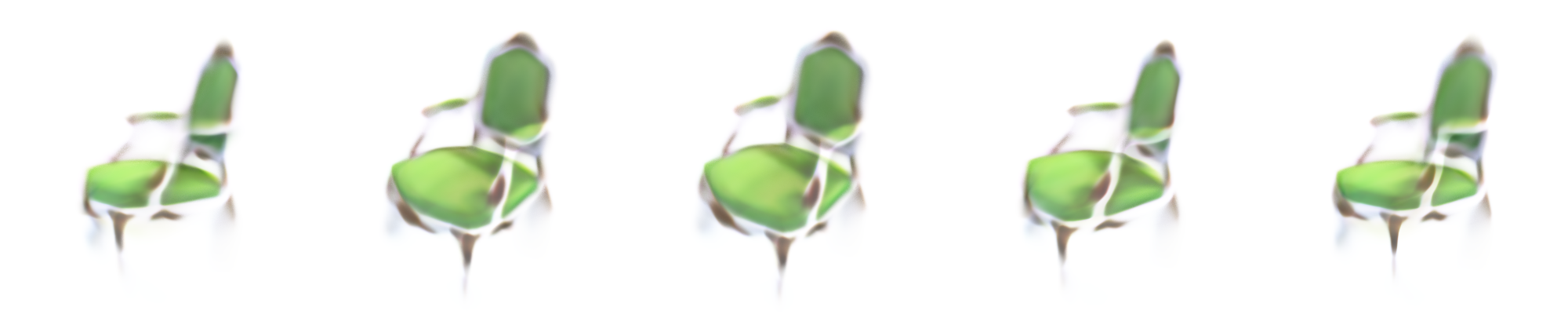} &
    \includegraphics[width=\wifig]{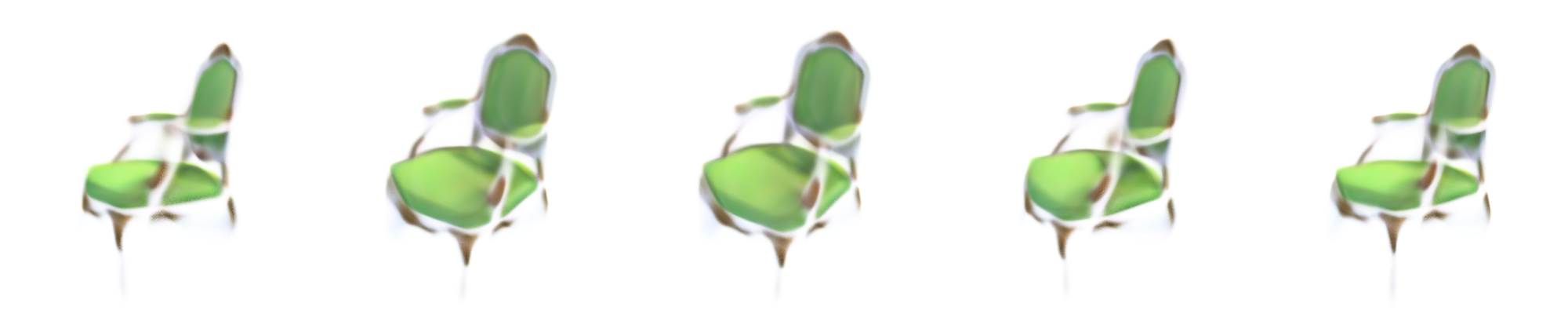} \\
    \includegraphics[width=\wifig]{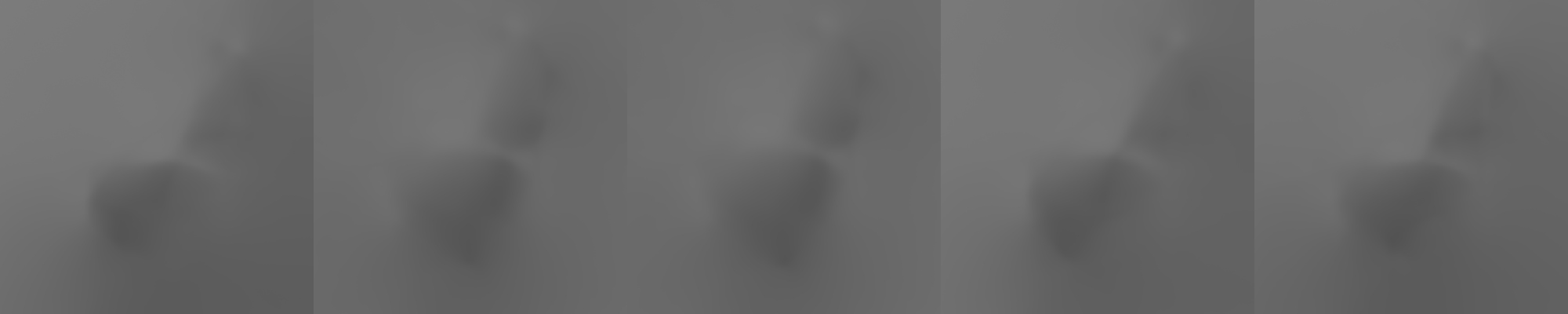} &
    \includegraphics[width=\wifig]{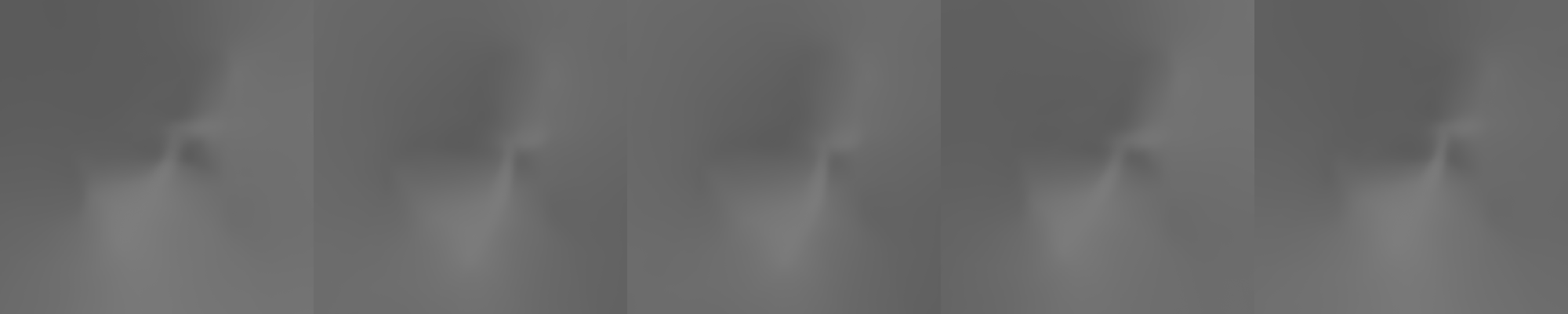} \\
        24.39 dB, \ang{27.18} & 25.10 dB, \ang{3.43}\\
     \midrule
    \includegraphics[width=\wifig]{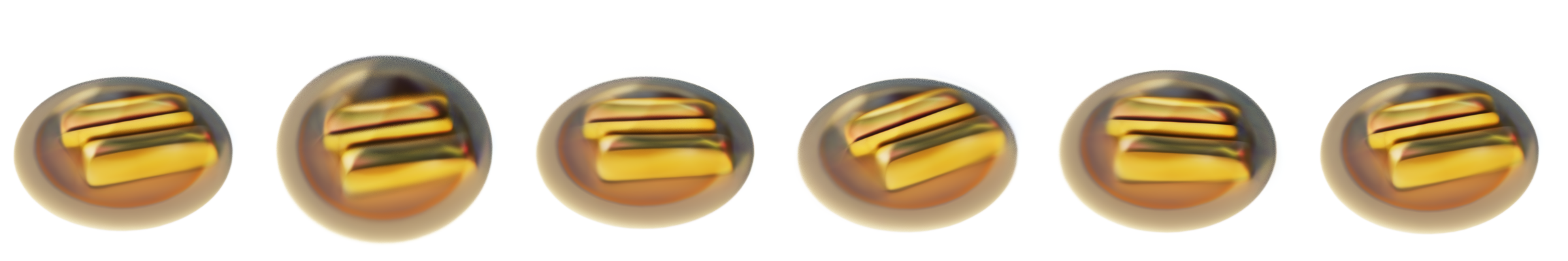} &
    \includegraphics[width=\wifig]{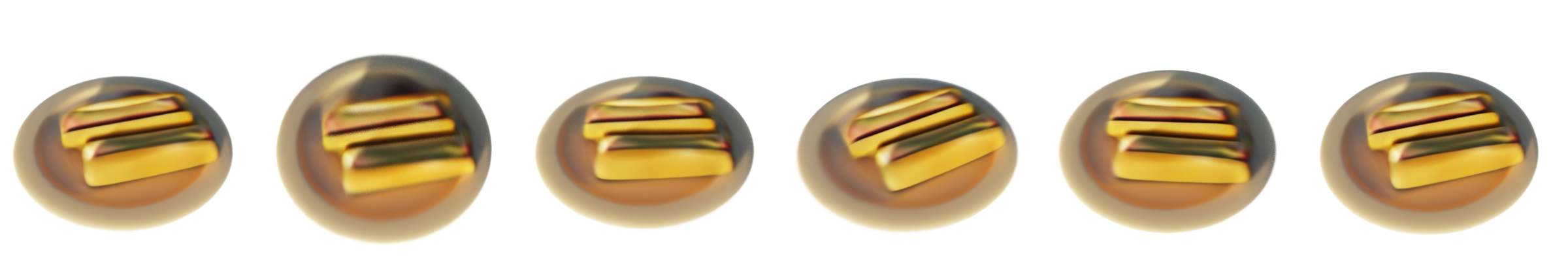} \\
    \includegraphics[width=\wifig]{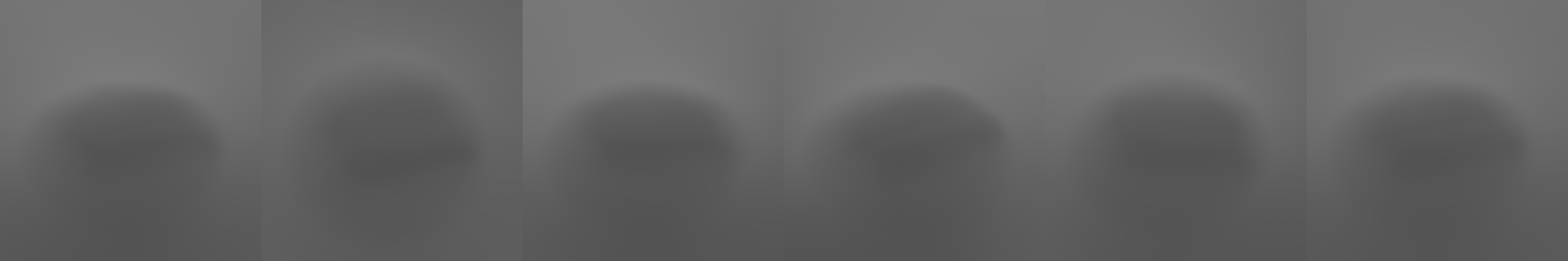} &
    \includegraphics[width=\wifig]{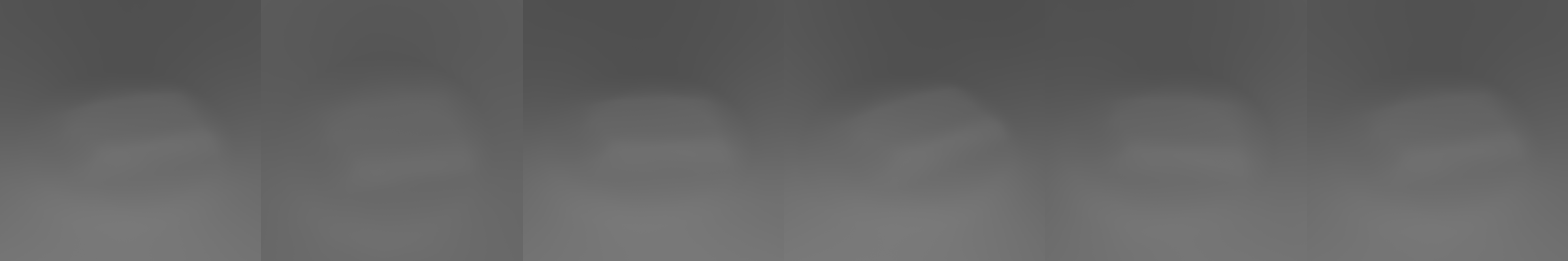} \\
    27.49 dB, \ang{18.37} & 27.73 dB, \ang{0.37} \\
    \midrule
    \includegraphics[width=\wifig]{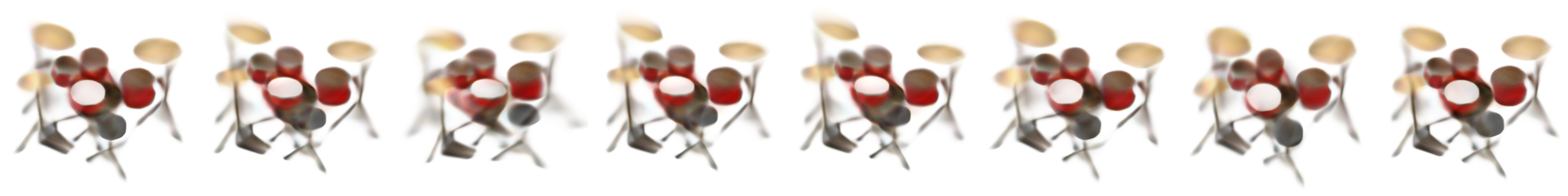} &
    \includegraphics[width=\wifig]{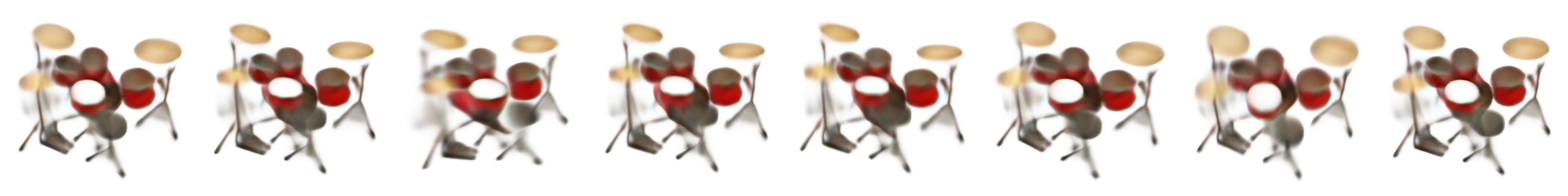} \\
    \includegraphics[width=\wifig]{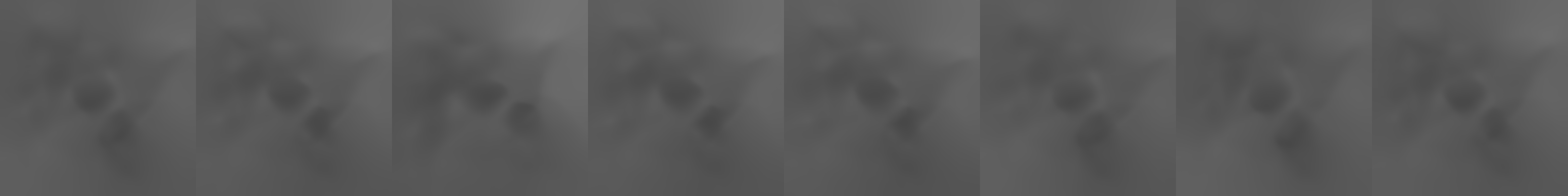} &
    \includegraphics[width=\wifig]{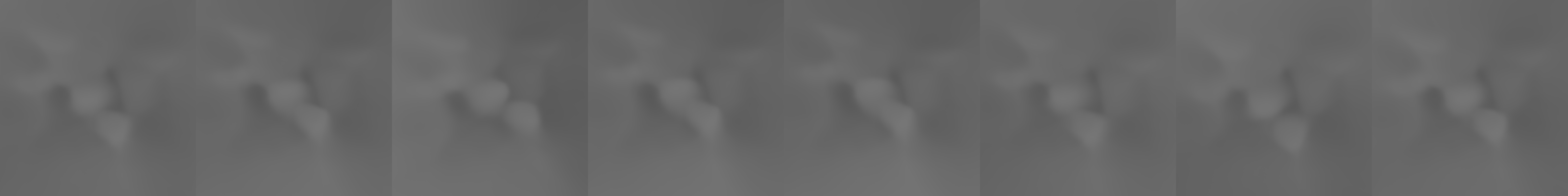} \\
    19.09 dB, \ang{16.89} & 19.66 dB, \ang{1.33} \\
    \midrule
    \includegraphics[width=\wifig]{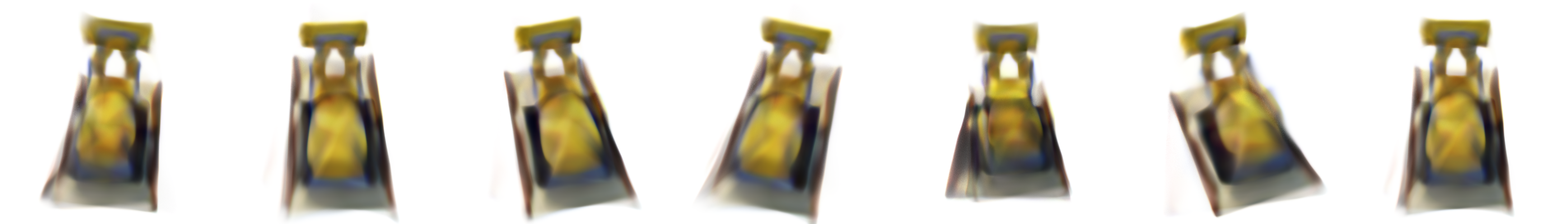} &
    \includegraphics[width=\wifig]{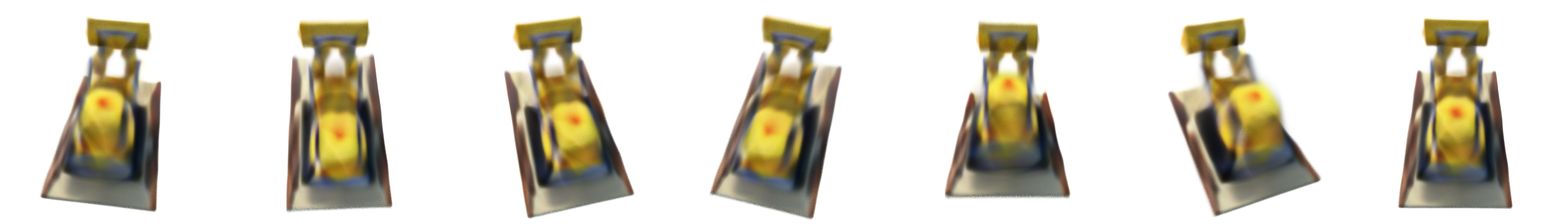} \\
    \includegraphics[width=\wifig]{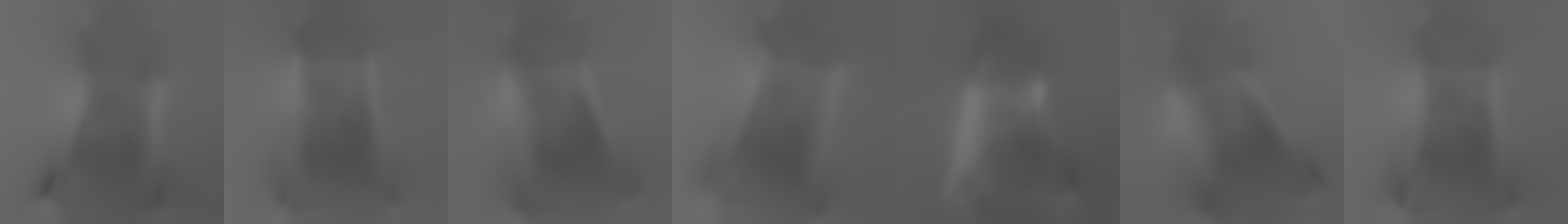} &
    \includegraphics[width=\wifig]{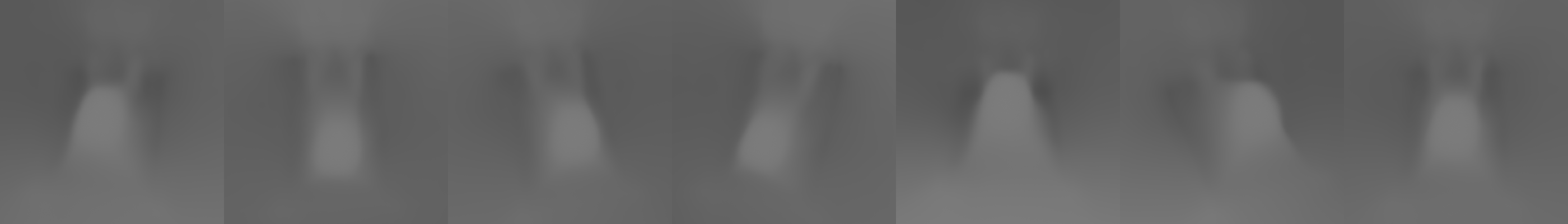} \\
    19.98 dB, \ang{20.81} & 21.74 dB, \ang{0.42} \\
    \midrule
    \includegraphics[width=\wifig]{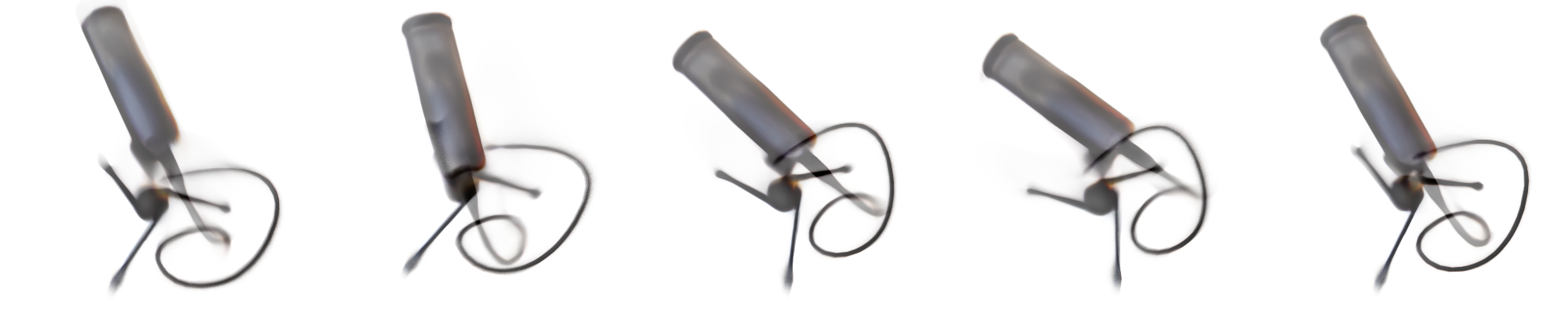} &
    \includegraphics[width=\wifig]{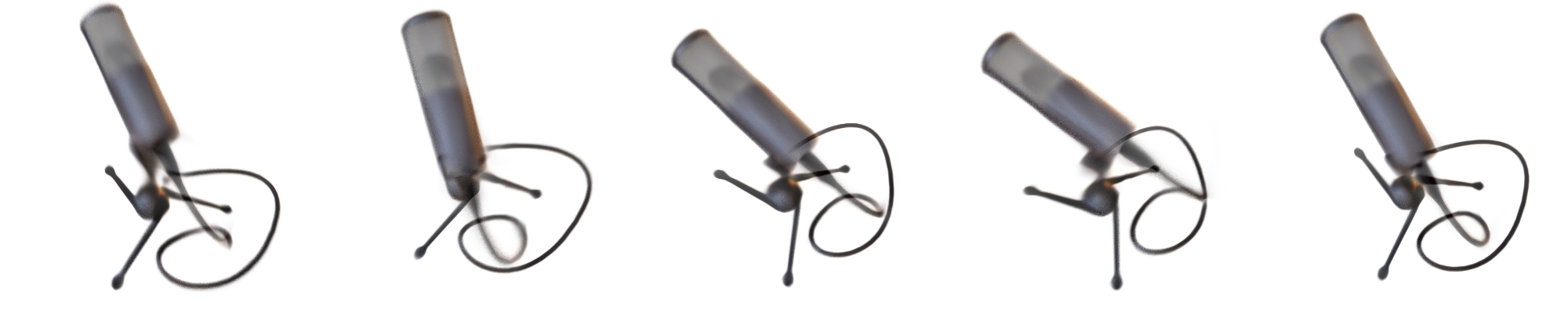} \\
    \includegraphics[width=\wifig]{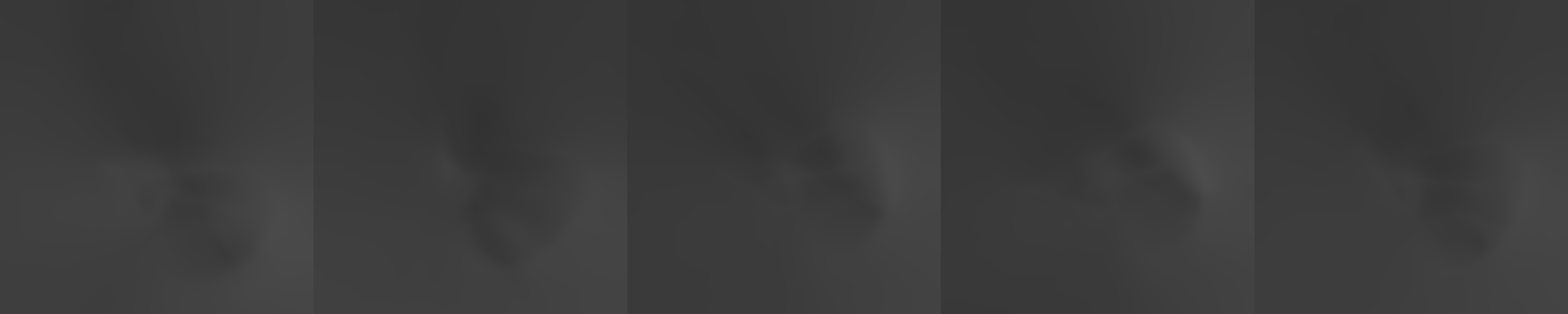} &
    \includegraphics[width=\wifig]{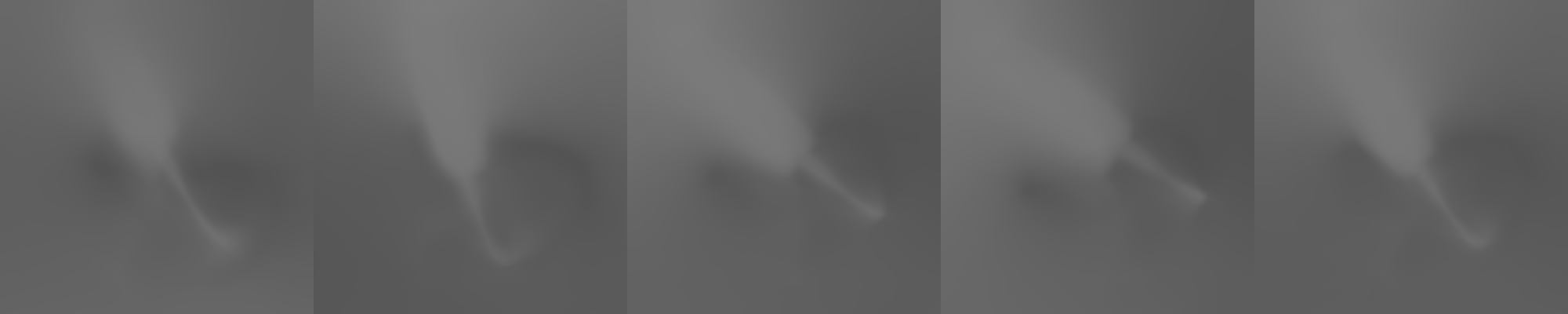} \\
    23.00 dB, \ang{21.06} & 24.32 dB, \ang{2.57} \\
    \midrule
    \small \textbf{w/o ambiguity resolution} & \textbf{w/ambiguity resolution}\\
    \midrule
    \end{tabular}
    \caption{\textbf{Mirror symmetry ambiguity.} For specific mini-scenes, we present renderings, disparity maps, PSNRs between the renderings and the groundtruth, and relative rotation errors (\textit{lower is better}) for LU-NeRF with and without the proposed solution to the mirror-symmetry ambiguity. 
    Brightness is inversely related to depth in the disparity map. The groundtruth depth maps are not available with the dataset.}
    \vspace{-4mm}
    \label{fig:mini-scenes}
\end{figure}

\noindent
\textbf{Mirror symmetry ambiguity.} 
Tab.~\ref{tab:mirror} shows the performance of our full method with and without the proposed solution to the mirror-symmetry ambiguity (Sec.~\ref{sec:mirror-symmtry}).  Resolving the ambiguity improves performance consistently, confirming the importance of this component to our pipeline.
For closer inspection, we present qualitative results for LU-NeRF with and without ambiguity resolution for select mini-scenes in Fig.~\ref{fig:mini-scenes}.  
Fig.~\ref{fig:mini-scenes} presents a visual comparison between LU-NeRF with and without the proposed solution to the mirror-symmetry ambiguity.  Without the ambiguity resolution, the predicted depths are reflected across a plane parallel to the image plane (having the effect of inverted disparity maps), and the poses are reflected across the center camera of a mini-scene. Our LU-NeRF-2 rectifies the predicted geometry and local camera poses, which effectively resolves the ambiguity.

\begin{table}
\centering    
\resizebox{\columnwidth}{!}{
  \begin{tabular}{c c c c c c c}
  \toprule
  Ambiguity & Chair & Hotdog & Lego & Mic & Drums \\
  \midrule
  w/o resolution & 39.14 & 138.9 & 0.48 & 107.9 & 11.35\\
  w/ resolution & \textbf{4.24} & \textbf{0.23} & \textbf{0.07} & \textbf{0.84} & \textbf{0.05} \\
  \bottomrule
  \end{tabular}
}
\caption{\textbf{Mirror symmetry ambiguity.}
  The mean rotation error in degrees for our pipeline (starting with the optimal graph), with and without the proposed strategy to resolve the ambiguity.}
  \label{tab:mirror}
\end{table}

\section{Discussion}
\label{sec:discussion}

In this work, we propose to estimate the neural scene representation and camera poses jointly from an unposed image collection through a process of synchronizing local unposed NeRFs. 
Unlike prior works, our method does not rely on a proper prior pose distribution and is flexible enough to operate in general \SE(3) pose settings. 
Our framework works reliably in low-texture or low-resolution images and thus complements the feature-based SfM algorithms.  Our pipeline also naturally exploits sequential image data, which is easy to acquire in practice.

One limitation of our method is the computational cost, which can be relieved by recent advances in neural rendering~\cite{advances_in_neural_rendering}.
Another limitation is the difficulty in building graphs for unordered scenes, which is a promising direction for future work. 

\section{Acknowledgements}
We thank Zhengqi Li and Mehdi S. M. Sajjadi for fruitful discussions.
The research is supported in part by NSF grants \#1749833 and \#1908669. Our experiments were partially performed on the University of Massachusetts GPU cluster funded by the Mass. Technology Collaborative.

{\small
\bibliographystyle{ieee_fullname}
\bibliography{egbib}
}

\clearpage
\appendix
\noindent{\Large \textbf{Appendix}}\label{sec:appendix}

\section{Experimental details}

\subsection{Training GNeRF}
We trained GNeRF~\cite{meng2021gnerf} using the official codebase\footnote{\url{https://github.com/quan-meng/gnerf}}.
The scores of GNeRF in our experiments are overall better than those reported in VMRF~\cite{vmrf},
likely because we trained GNeRF for more iterations. In our experiments, we train GNeRF for 60K iterations which take 48 hours on a single NVIDIA GeForce GTX 1080Ti. 
We notice that GNeRF is prone to mode collapse in the adversarial training stage, \ie, the generator produces the same or similar sets of outputs with negligible variety, which is a well-known issue for GAN-based models~\cite{arjovsky2017wasserstein}. 
To achieve similar performance reported in GNeRF and VMRF, we train 5 GNeRF models per prior pose setting and report the results from the best one selected according to the performance on the validation set.
Specifically, $35\%$ of the training trials (26 out of 75) suffered from the mode collapse issue on the unordered image collections. 

\begin{table*}[th!]
\resizebox{\textwidth}{!}{
\centering
\begin{tabular}{@{}lccccccccccccccccccc@{}}
  \toprule
                & \multicolumn{3}{c}{Chair} &               & \multicolumn{3}{c}{Hotdog} &  & \multicolumn{3}{c}{Drums} &               & \multicolumn{3}{c}{Lego} &  & \multicolumn{3}{c}{Mic}                                                                                                                                \\
  \cmidrule{2-4}\cmidrule{6-8}\cmidrule{10-12}\cmidrule{14-16}\cmidrule{18-20}
                & PSNR $\uparrow$                  & SSIM $\uparrow$      & LPIPS $\downarrow$                  &  & PSNR $\uparrow$                  & SSIM $\uparrow$      & LPIPS $\downarrow$                &  & PSNR $\uparrow$       & SSIM $\uparrow$      & LPIPS $\downarrow$     &  & PSNR $\uparrow$       & SSIM $\uparrow$      & LPIPS $\downarrow$     &  & PSNR $\uparrow$       & SSIM $\uparrow$      & LPIPS $\downarrow$        \\
  \midrule
GNeRF \ang{90}  & 31.30                     & 0.95          & 0.08                       &  & 32.00                     & 0.96          & 0.07                     &  & 24.30          & 0.90          & 0.13          &  & 28.52          & 0.91          & 0.96          &  & 31.07          & 0.09          & 0.06          \\
GNeRF \ang{120} & 28.31                     & 0.92          & 0.12                       &  & 25.91                     & 0.92          & 0.15                     &  & 22.04          & 0.88          & 0.19          &  & 23.10          & 0.86          & 0.95          &  & 25.98          & 0.16          & 0.08          \\
GNeRF \ang{150} & 22.63                     & 0.88          & 0.22                       &  & 23.03                     & 0.90          & 0.24                     &  & 20.11          & 0.87          & 0.21          &  & 22.02          & 0.85          & 0.93          &  & 22.71          & 0.18          & 0.12          \\
\midrule
GNeRF \ang{180} (2DOF) & 21.60                     & 0.87          & 0.25                       &  & 24.57                     & 0.92          & 0.18                     &  & 18.94 & 0.84 & 0.30          &  & 20.48          & 0.85          & 0.20 &  & \textbf{23.80} & \textbf{0.94}          & 0.12          \\
Ours (3DOF)     & \textbf{30.57}            & \textbf{0.95} & \textbf{0.05}              &  & \textbf{34.55}            & \textbf{0.97} & \textbf{0.03}            &  & \textbf{23.53}          & \textbf{0.89}   & \textbf{0.12} &  & \textbf{28.29} & \textbf{0.92} & \textbf{0.06}      &  & 22.58          & 0.91 & \textbf{0.08} \\
  \bottomrule
\end{tabular}
}
\caption{\textbf{Novel view synthesis on unordered image collection.} We trained the GNeRF with the publicly available code.
  Each GNeRF variation is described by the assumed elevation range.
  GNeRF \ang{180} is the closest to our method but still has only 2 degrees of freedom for assuming upright cameras.
  Our method outperforms the unconstrained GNeRF while being more general for considering
  arbitrary rotations with 6 degrees-of-freedom.
  }
\label{tab:view-syntheis-v2}
\end{table*}

\subsection{Test-time optimization for view synthesis}\label{sec-testtimeopt}
Following the procedure established by prior works~\cite{lin2021barf,meng2021gnerf} for evaluating novel view synthesis, we register the cameras of the test images using the transformation matrix obtained via Procrustes analysis on the predicted and groundtruth cameras of the training images;
we then optimize the camera poses of the test images with the photometric loss to factor out the effect of the pose error on the view synthesis quality.
In the GNeRF pipeline, the test-time optimization is interleaved with the training process, so the total number of iterations depends on the number of training steps seen for the best checkpoint. In our experiments, GNeRF runs approximately 4230 test-time optimization iterations.

\subsection{Qualitative and quantitative comparisons}

In the main text, we quote the scores from VMRF where the results on Hotdog are missing. Here we also train GNeRF using the official codebase and report the results in Table~\ref{tab:view-syntheis-v2}. This allowed us to generate GNeRF results on the Hotdog scene, and observe the mode collapse reported in the previous section.

Overall, our method outperforms both GNeRF and VMRF under an unconstrained pose distribution,
while also being more general --  
our method works with arbitrary 6 degrees-of-freedom rotations (6DOF), 
while the baselines assume upright cameras (2DOF) on the sphere, even when the range of elevations is not constrained.

\section{Supplementary analysis}

\subsection{Pose synchronization and refinement}\label{sec:sync_refine}
Table~\ref{tab:global-pose} demonstrates the necessity of our pose synchronization and refinement steps. The pose synchronization aggregates the local pose estimation from LU-NeRF and provides a rough global pose configuration, and the pose refinement step further improves the global poses.  

\begin{table}[]
\resizebox{\columnwidth}{!}{
\centering
\begin{tabular}{c cc cc}
 &  \multicolumn{2}{c}{Rotation Error [\textdegree]} &  \multicolumn{2}{c}{Translation Error}  \\
\multirow{1}{*}{Scenes} & Pose Sync. & Pose Refine. & Pose Sync. & Pose Refine. \\
\cmidrule(l{10pt}r{10pt}){2-3} \cmidrule(l{10pt}r{10pt}){4-5}
Lego & 16.50 & 0.07 & 0.85 & 0.00 \\
Chair & 20.53 & 4.24 & 1.08 & 0.16 \\
Hotdog & 21.06 & 0.23 & 1.17 & 0.01 \\
Drums & 14.30 & 0.05 & 0.86 & 0.00 \\
Mic & 35.48 & 0.84 & 1.90 & 0.02 \\
\midrule
Mean & 21.57 & 1.09 & 1.17 & 0.04 
\end{tabular}
}
\caption{\textbf{Pose synchronization and refinement.} The pose synchronization step provides a rough global pose configuration, and all camera poses are further optimized during the pose refinement step. We use unordered image collections in this experiment.}
\label{tab:global-pose}
\end{table}

\subsection{Case study on Drums}\label{sec:failure}
\begin{figure}[h!]
\centering
\includegraphics[width=0.5\linewidth]{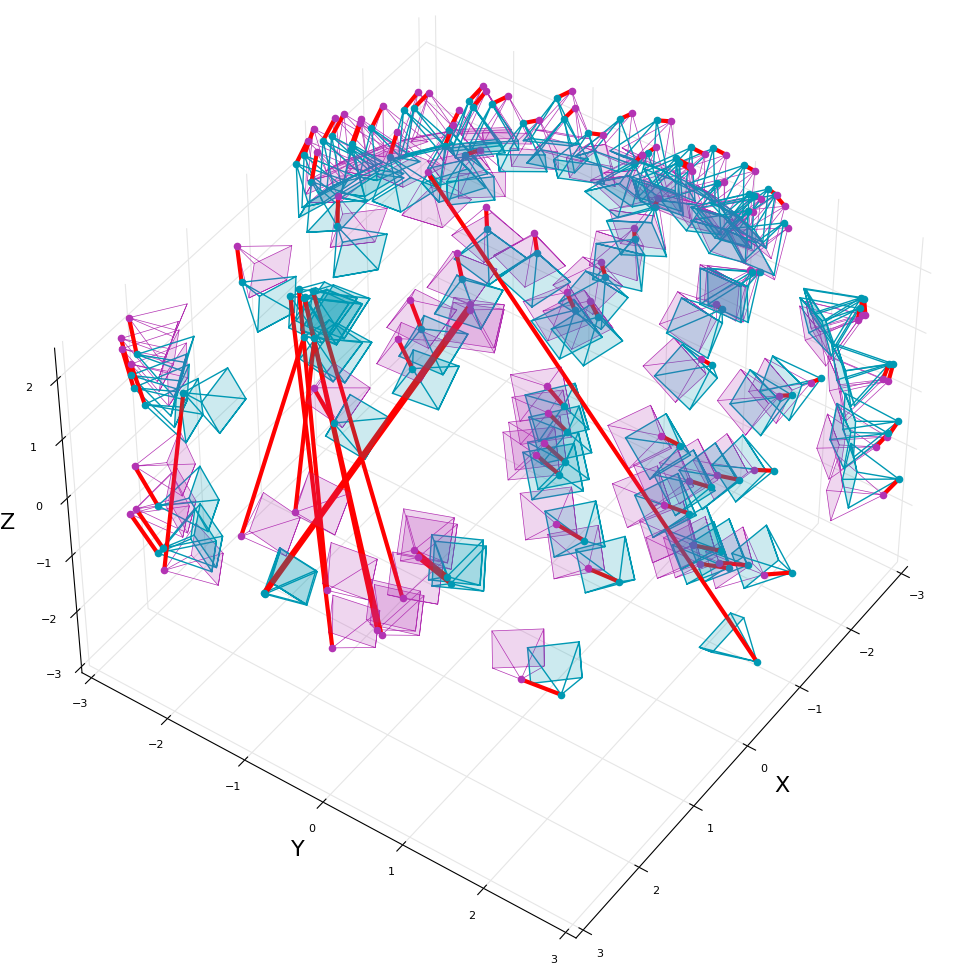}
\caption{\textbf{Camera pose predictions on Drums. 
}}
\label{fig:drums}
\end{figure}

In this section, we take a closer look at the performance of our model on Drums, the worst-performing scene in Table~\ref{tab:view-syntheis-v2}.  The mean rotation error over the 100 cameras is 12.39$^\circ$ (see Table 1 of the main paper). Figure~\ref{fig:drums} shows the estimated camera poses juxtaposed with the ground truth cameras after Procrustes alignment. We can see that there is a small cluster of poorly posed images. Since Procrustes finds the optimal least-squares global alignment between predicted and true camera poses, it is severely affected by these outlier images.  
A subtle consequence of this is that the test time optimization, described in Sec.~\ref{sec-testtimeopt}, may not be sufficient to evaluate the novel view synthesis results accurately and quantitatively. 
Due to the exaggerated misalignment from Procrustes in Drums, we may need to increase the number of iterations in order to converge to a more accurate viewpoint.  
Indeed, we find simply increasing the number of test-time optimization iterations from 100 to 1000 dramatically improves the rendering metrics:  PSNR increases from 14.26 to 23.53, SSIM increases from 0.71 to 0.89, and LPIPS decreases from 0.30 to 0.12.

\subsection{Effect of depth regularization}\label{sec:ablation}

Similar to RegNeRF~\cite{Niemeyer2021Regnerf}, we encourage the smoothness of the predicted depth maps and apply a depth regularization on small patches.
We sample patches rendered from the cameras whose poses are jointly optimized with the 3D representation, while RegNeRF uses groundtruth poses for the observed views and samples the patches from unobserved viewpoints.
We find that such depth regularization is crucial to the success of LU-NeRF.
Table~\ref{tab:depth} shows that
incorporating the depth regularization significantly improves the pose estimation accuracy of LU-NeRF -- the median relative rotation errors decrease from \ang{11.08} to \ang{4.69} while the mean drops from \ang{11.93} to \ang{8.26}. Even though the maximum relative rotation error is smaller without the depth regularization, the Shonan averaging~\cite{dellaert2020shonan} fails to converge to a reasonable global pose configuration.  

\begin{table}[t]
\resizebox{\columnwidth}{!}{
\centering
{
\begin{tabular}{c rrr rrr}
 \multirow{3}{*}{Scenes} & \multicolumn{3}{c}{LU-NeRF} & \multicolumn{3}{c}{LU-NeRF}   \\
 & \multicolumn{3}{c}{w/o depth regularization} & \multicolumn{3}{c}{w/ depth regularization} \\
 \cmidrule(l{10pt}r{10pt}){2-4} \cmidrule(l{10pt}r{10pt}){5-7}
  & \multicolumn{1}{c}{Mean} & \multicolumn{1}{c}{Median} & \multicolumn{1}{c}{Max} & \multicolumn{1}{c}{Mean} & \multicolumn{1}{c}{Median} & \multicolumn{1}{c}{Max}\\ 
Chair & 14.18 & 13.33 & 38.26  & \textbf{9.41} & \textbf{4.89} & \textbf{33.05}  \\
Hotdog &  10.75 & 9.49 & \textbf{29.41}  & \textbf{10.69} & \textbf{7.77} & 33.29 \\
Lego & 10.50 & 10.08 & \textbf{28.27}  & \textbf{5.58} & \textbf{1.33} & 30.88 \\
Mic & 12.88 & 11.70 & \textbf{25.99}  & \textbf{10.27} & \textbf{7.05} & 30.03 \\
Drum  &  11.32 & 10.44 & \textbf{24.10} & \textbf{5.37} & \textbf{2.40} & 29.27 \\
\midrule
Mean & 11.93 & 11.08 &	\textbf{29.21}  & \textbf{8.26} & \textbf{4.69} & 31.30 
\end{tabular}
}
}
\caption{\textbf{Effect of depth regularization on the pose estimation.}
  We report the mean relative rotation error (\textdegree) with and without depth regularization. The relative rotations are computed between the center camera and its neighbors within a mini-scene. The relative rotation error (\textit{lower is better}) is defined as the rotation angle between the predicted relative rotations and the groundtruth.}
\label{tab:depth}
\end{table}

\subsection{Computational cost}
Table~\ref{table:time-cost} presents the computational cost of the proposed framework.
We randomly sample 30 mini-scenes and report the average training time for LU-NeRF-1 and LU-NeRF-2. The LU-NeRFs for different mini-scenes are independent and thus can be trained in parallel. 
The training of LU-NeRFs and the global pose refinement with BARF~\cite{lin2021barf} can be significantly accelerated with some recent advances in learnable scene representations (\eg PlenOctrees~\cite{yu2021plenoctrees}, InstantNGP~\cite{mueller2022instantngp}).

\begin{table}[!htbp]
  \centering
  \begin{tabular}{c c}
  \toprule
  Stage & Running time \\
  \midrule
  LU-NeRF-1 & 1.08 hours \\
  LU-NeRF-2 & 0.89 hours \\
  Pose synchronization & 3.18 seconds \\
  Pose refinement & 4.40 hours \\
  \bottomrule
  \end{tabular}
  \caption{\textbf{Computational cost.}
    We report the mean time for training a single LU-NeRF-1, a single LU-NeRF-2, and the final refinement step on an NVIDIA Tesla P100.
  The pose synchronization step runs on CPU and has a negligible running time.}
  \label{table:time-cost}
\end{table}

\section{Implementation details}

\subsection{Building connected graphs}\label{sec:graph}

Given a distance function $\text{dist}(\cdot, \cdot)$, image descriptors $\{f_i\}^N$ and corresponding cameras $\{C_i\}^N$ where $i \in \{0, \dots, N-1\}$ is the image index and $N$ is the total number of images, as the first step, we build a minimal spanning tree (MST) using Kruskal's algorithm.
Each node on the MST represents a camera and the weight $W_{ij}$ on the edge that connects camera $C_i$ and $C_j$ is the distance $\text{dist}(f_i, f_j)$ between the descriptors of image $I_i$ and $I_j$.
To ensure each mini-scene contains at least $K$ images ($K=5$ by default), we augment the MST by adding nearest neighbors for nodes that have less than $K-1$ connected nodes in the MST.
We also ensure that each edge appears in both mini-scenes centered at the endpoints of the edge, such that there are at least two measurements for the relative pose between two connected cameras. Having multiple measurements allows for estimating the confidence of the predicted relative poses and identifying LU-NeRF failures (see Sec.~\ref{sec:implementation}). 

\subsubsection{Distance functions}

\noindent
\textbf{Motivation.} We intentionally experiment with simplistic approaches to compute image similarity in our graph-building procedure since our primary contribution in this work is the local-to-global pose and scene estimation starting with LU-NeRF on mini-scenes. In practice, depending on the application context, there are likely different cues or weak supervision that can be exploited for graph building (as we do for ordered sequences). We leave it to future work to explore more sophisticated unsupervised/self-supervised techniques for building neighborhood graphs.

In our experiments, we tried two different features to build the connected graph: self-supervised DINO features~\cite{caron2021emerging} and raw RGB values with $\ell_1$ distance. 

\noindent
\textbf{DINO features.}
We extract semantic object parts by applying K-means clustering on the image collections~\cite{amir2021deep}.
The number of parts is 10 by default in our experiments.
We build an image descriptor in a similar way as the Histogram-of-Gradients (HoG).
Specifically, we evenly split the part segmentation maps into $4\times 4$ grid and compute a part histogram within each cell. We then normalize the histogram per cell into a unit vector and use the concatenated $16$ histograms as the image descriptors.
The cosine distance between these descriptors is used to build the MST and final graph.

\noindent
\textbf{$\ell_1$ on RGB values.}
We estimate the distance between two images as the minimum $\ell_1$ cross-correlation
over a set of rotations and translations.
Formally, we compute
\begin{align}
  \text{dist}(I_1, I_2) &=  \argmin_{t,\theta} \sum_{x}\mid I_1(x) - I_2(R_\theta x+t) \mid,
\end{align}
where $x$ is the pixel index, $t$ is in a $5\times 5$ window centered at 0,
$R_\theta$ is an in-plane rotation by $\theta$ and $\theta \in \{0, \pi\}$.
When the minimum is found at $\theta \neq 0$, we augment the input scene with the in-plane rotated image and
correct its estimated pose by $\theta$.

We found that considering the in-plane rotations here is useful because of symmetries --
some scenes of symmetric objects can contain similar images from cameras separated by large
in-plane rotations.
This is problematic for LU-NeRFs because they initialize all poses at identity.
Augmenting the scene with the closest in-plane rotations makes the target poses closer to
identity and helps convergence.

\noindent
\textbf{Metric selection}
In experiments with unordered image collections, we used the $\ell_1/\text{RGB}$ metric for Lego and Drums,
and the DINO metric for Chair, Hotdog, and Mic. The RGB metric fails to build useful graphs for Hotdog, Mic and Ship,
while the DINO metric fails for Lego, Drums, and Ship.
No graph-building step is necessary on ordered image collections since the order determines the graph.

\subsubsection{Analysis}
\begin{figure}
\centering
\includegraphics[width=\linewidth]{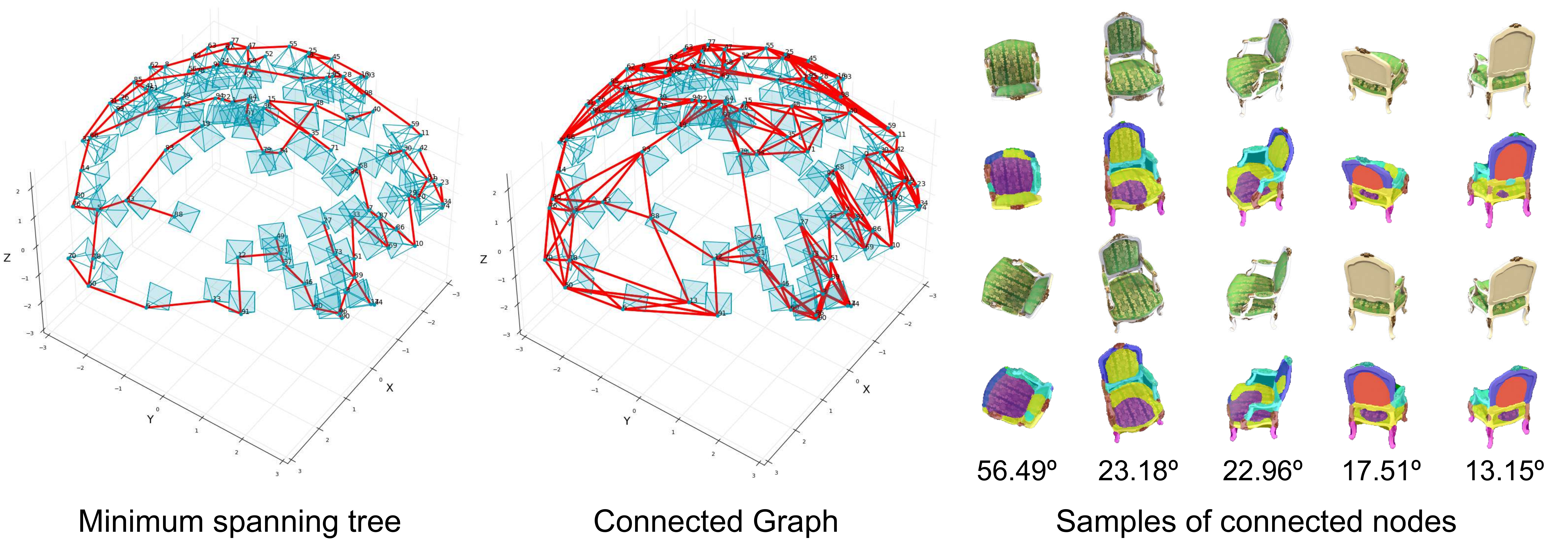}
\caption{\textbf{Graph built with DINO features on Chair.} The minimum spanning tree (left), the connected graph (middle), and samples of connected image pairs (right). In the right panel, each column presents two images that are connected on the graph (1st and 3rd row), the corresponding part co-segmentation maps~\cite{amir2021deep} (2nd and 4th row), and rotation distance between the two views (bottom).}
\label{fig:graph-dino}
\end{figure}

Figure~\ref{fig:graph-dino} presents the MST, the connected graph, and image pairs that are connected in the graph on the Chair scene from the NeRF-synthetic dataset when using the DINO features.
Surprisingly, the self-supervised ViT generalizes well on unseen objects and the learned representations are robust to  viewpoint change (see the last column of Figure~\ref{fig:graph-dino}).

\begin{figure}
\centering
\includegraphics[width=\linewidth]{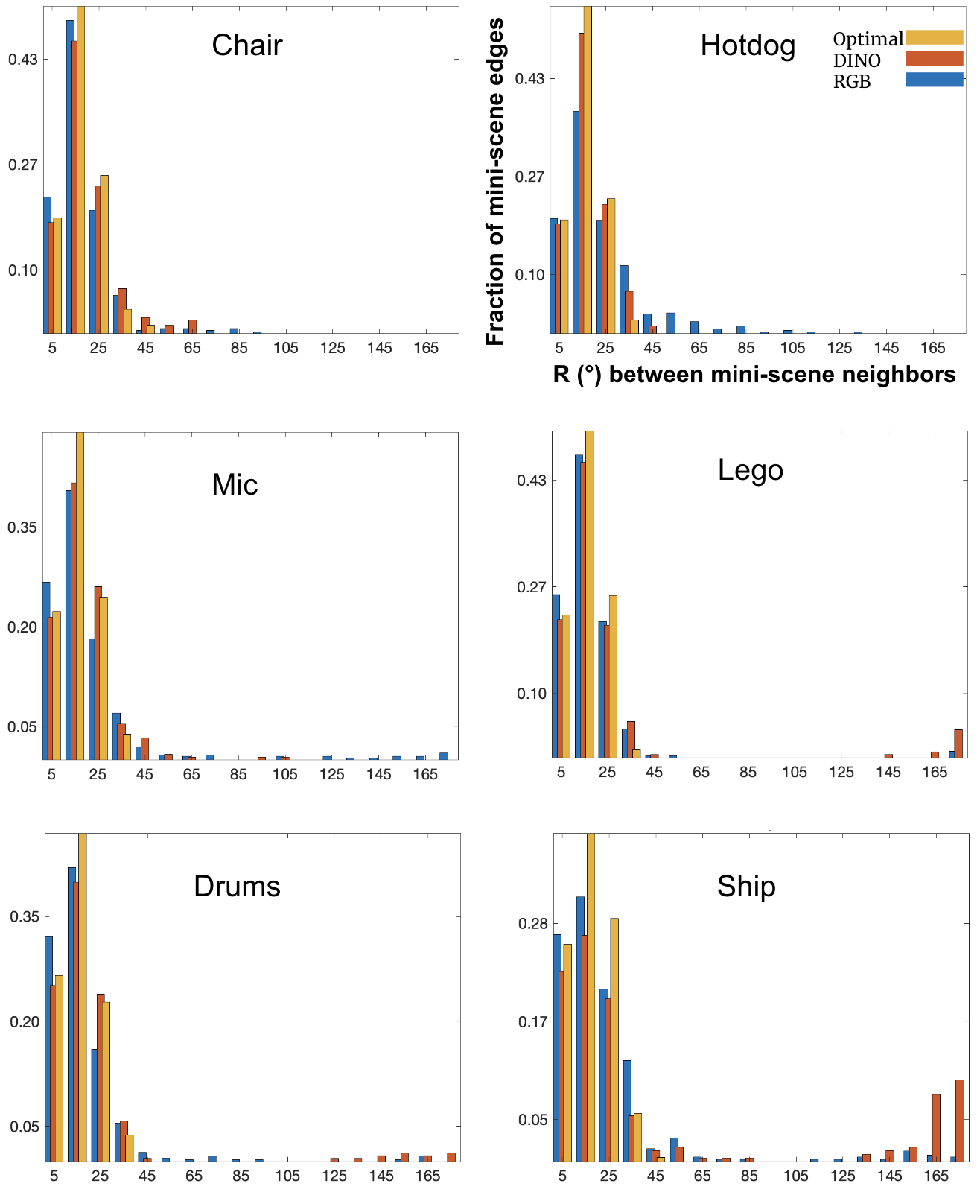}
\caption{\textbf{Graph statistics.} We compare the rotation distance between mini-scene neighbors on the optimal graph built with groundtruth camera poses, the graph built with DINO features, and the one built with RGB features.  For most scenes both DINO and RGB mini-scenes include outlier images (close to 180$^\circ$ distance) which our pipeline needs to deal with.}
\label{fig:graph-distance}
\end{figure}

Figure~\ref{fig:graph-distance} presents an analysis of the connected graphs built with DINO and RGB features. Both features provide outlier-free connected graphs on Chair. The graphs built with DINO contain much fewer outliers on Hotdog and Mic, while RGB features induce clearer graphs on Drums and Lego. Both DINO and RGB features produce more outliers on Ship than other scenes.

\noindent
\textbf{Optimal graph vs noisy graph.}
To analyze the effect of graph building on the unordered image collection,
we build an \textit{oracle} outlier-free connected graph with groundtruth camera poses.
Table~\ref{tab:graph} compares the performance of our method with the optimal graphs and noisy graphs built with DINO/RGB features. 
Outliers in the connected graph may hurt the performance of LU-NeRF. 
Nevertheless, with our simple graph-building methods based on DINO/RGB features,
our method outperforms the baselines when they are not given prior constraints on the camera pose distributions. 

We notice that the performance with the optimal graph is worse than that with the noisy graph on Chair.
The ``optimal'' graph minimizes camera distances, but it is not guaranteed to be the best choice for LU-NeRF.
\eg, issues like mirror symmetry ambiguity (Sec.~\ref{sec:mirror-symmtry}) can arise more often when cameras are in close proximity, and there is randomness inherent in training neural networks.

\begin{table}[h!]
\resizebox{\columnwidth}{!}{
\centering
\begin{tabular}{ccccc}
\toprule
&  \multicolumn{2}{c}{Rotation Error [\textdegree]} &  \multicolumn{2}{c}{Translation Error}  \\
\multirow{1}{*}{Scenes}  & Optimal graph & Noisy graph & Optimal graph & Noisy graph \\
\cmidrule(l{10pt}r{10pt}){2-3} \cmidrule(l{10pt}r{10pt}){4-5}
Chair & 4.24 & 2.64 & 0.16 & 0.09 \\
Hotdog & 0.23 & 0.24 & 0.01 & 0.01  \\
Lego & 0.07 &  0.09 & 0.00 & 0.00 \\
Mic & 0.84  &  6.68 & 0.02 & 0.10 \\
Drums &  0.05  & 12.39 & 0.00 & 0.23\\
\bottomrule
\end{tabular}
}
\caption{\textbf{Optimal graphs vs noisy graphs.} The outliers in the noisy connected graph built with DINO/RGB features may hurt the performance of our method in camera pose estimation. The clean graph is built from the ground-truth camera poses.}
\label{tab:graph}
\end{table}

\subsection{LU-NeRF architecture and training details}\label{sec:implementation}

In the training of LU-NeRF, we do not apply the coarse-to-fine strategy proposed in BARF~\cite{lin2021barf}; we sample 2048 rays per mini-batch and adopt the learning rate schedule for pose and MLP parameters from BARF; we remove the positional encoding and view-dependency, 
and use a compact 4-layer MLP to reduce the memory cost and speed up the training. 
We set the initial camera poses to identity. 
We have experimented with random initialization around identity but observed no significant difference.
We terminate the training of LU-NeRF-1 if the average change of the camera rotations in a mini-scene is less than \ang{0.125} within 5k iterations.
We train LU-NeRF-2 for 5k iterations with frozen initial poses and then jointly optimize camera poses and neural fields for 45k iterations. 
We apply depth regularization on small patches ($2\times 2$ by default) in both LU-NeRF-1 and LU-NeRF-2.

\subsection{Synchronization and refinement details}
In the pose synchronization step, we apply the off-the-shelf Shonan averaging\footnote{\url{https://github.com/dellaert/ShonanAveraging}}~\cite{dellaert2020shonan},
which solves a convex relaxation of the problem described in Eqn. (1) of the main text,
while iteratively converting it to higher dimensional special-orthogonal spaces \SO(n) until it converges. We then optimize the translation with fixed camera rotation.

The input to the Shonan averaging is the relative pose estimations from LU-NeRF and the confidence of these pose estimations. 
Each pair of cameras may have multiple measures of the relative poses, as each camera may appear in multiple mini-scenes. 
We apply a simple heuristic to pick one measure from these multiple candidates: 
given two cameras $C_i$ and $C_j$ and their renderings $I_i$ and $I_j$, where $i < j$, if the PNSR of $I_i$ in the mini-scene centered at $C_i$ is higher than the PSNR of $I_i^\prime$ in the mini-scene centered at $C_j$, we use the pose estimation from the mini-scene $i$ as our relative pose estimation between camera $C_i$ and camera $C_j$.

To resolve the scale ambiguity across different mini-scenes, 
we first scale each mini-scene so that the MST edges connecting different mini-scenes are scale-consistent (MST construction is described in Sec.~\ref{sec:graph}). 
Specifically, we establish a reference scale by fixing it in one mini-scene and propagating it to others through the MST. 
We focus on edges linking mini-scene centers and rescale the mine-scenes so that overlapping edges share a consistent scale.

We then obtain the translation by solving a linear system $t_j - t_i = R_it_{ij}$ where $R_i$ is the rotation of camera $i$ from the Shonan method and $t_{ij}$ is the relative translation between camera $i$ and $j$ from LU-NeRF. 
Similar to the relative rotations, each pair of cameras may have multiple measures of the relative translation.
We use the same heuristic described above to pick one from the multiple measures. 
The translation optimization has also been implemented in the off-the-shelf Shonan averaging.

In the global pose refinement stage, we closely follow the default setting of BARF~\cite{lin2021barf} which jointly trains the MLP and refines the poses for 200k iterations with a coarse-to-fine positional encoding strategy. 

We utilize the camera visualization toolkit from BARF~\cite{lin2021barf} in our main paper and the Appendix.

\subsection{Dataset release and open sourcing.} We will release the newly ordered Blender sequences and open-source the code for our models.

For the sequential data sampled from the Objectron dataset~\cite{ahmadyan2021objectron}, Table~\ref{tab:objectron-baselines} reports the average rotation  and translation distance between all camera pairs as a reference for our quantitative evaluations in Table~6 in the main text.

\begin{table}[t]
\resizebox{\columnwidth}{!}{
\centering
\begin{tabular}{@{}
  l
  cccccc}
\toprule
                    & Bike          & Chair         & Cup           & Laptop        & Shoe          & Book          \\
\; Rotation         & 24.41        & 5.61          & 13.41             & 23.13             & 19.36          & 45.78             \\

\; Translation           & 4.09      & 1.20 & 0.63            & 1.79             & 0.82  & 1.43             \\
\bottomrule
\end{tabular}
}
\caption{\textbf{Camera baselines.} We report the average rotation [\textdegree] and translation distance between all camera pairs in the sequential data sampled from the Objectron datatset~\cite{ahmadyan2021objectron}.}
\label{tab:objectron-baselines}
\end{table}

\end{document}